\documentclass[runningheads]{llncs}

\usepackage{eccv}

\usepackage{eccvabbrv}

\usepackage{graphicx}
\usepackage{booktabs}
\usepackage{adjustbox}

\usepackage[accsupp]{axessibility}  
\usepackage{siunitx}

\usepackage{hyperref}

\usepackage{orcidlink}

\begin{document}

\title{On the Real-World Generalisability of \\ Optical Flow Models} 


\author{Petter Reijalt \orcidlink{0009-0008-2628-1636} \and
Sander Gielisse
\orcidlink{0009-0009-1224-2123} \and
Rickard Karlsson \orcidlink{0009-0002-9800-0836} \and
Jan van Gemert\orcidlink{0000-0002-3913-2786}}

\authorrunning{P.~Reijalt et al.}

\institute{
Computer Vision Lab, TU Delft, Delft, the Netherlands\\}

\maketitle

\begin{abstract}
Real-world deployment of vision models to broadly benefit society is arguably a main research objective. In optical flow, however, the difficulty to obtain the ground truth has focused research mainly on synthetic data and domain-specific benchmarks. Here, we investigate the severity of this mismatch. We study how well modern optical flow estimation models generalise to real-world video and question if accuracy on synthetic benchmark proxies actually predicts accuracy on real-world optical flow. To address this, we build a real-world evaluation benchmark and evaluate the real-world generalisability of a broad set of recent optical flow models using standard checkpoints. Our benchmark contains 8,204 frame pairs across TAP-Flow, Slow Flow, and our own dataset FlowFactor. FlowFactor is a manually annotated real-world benchmark of 1,000 HD frame pairs organised into four confounding factors: large displacements, repetitive textures, occlusions, and lighting variation. Each setting mainly varies only one factor, enabling diagnostic, confounder-specific analysis.  Using FlowFactor, we reveal that performance on varying lighting and large displacements correlates most strongly with real-world accuracy, and that improvements on large-motion regimes can trade off against robustness in small-motion, stationary scenes. Our experiments show that progress on Sintel, KITTI and Spring only weakly predicts accuracy on real-world data, highlighting the need for a broad real-world optical flow benchmark. Interestingly,  scaling up the amount of training data does not necessarily resolve the gap, calling for new innovative research instead of simply scaling data and compute. Code and datasets are available at \url{https://github.com/Petter6/real-world-optical-flow}. 
\end{abstract}

\begin{figure}
\begin{tabular}{ll@{\hskip 5mm}rr}
\multicolumn{2}{c}{Synthetic optical flow} & \multicolumn{2}{c}{Real-world optical flow} \\
\includegraphics[width=0.235\linewidth]{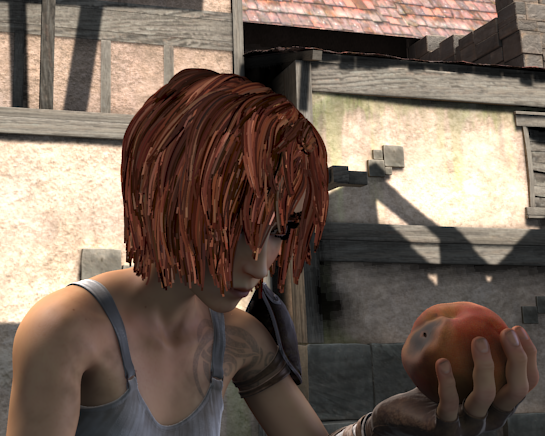} &
\includegraphics[width=0.235\linewidth]{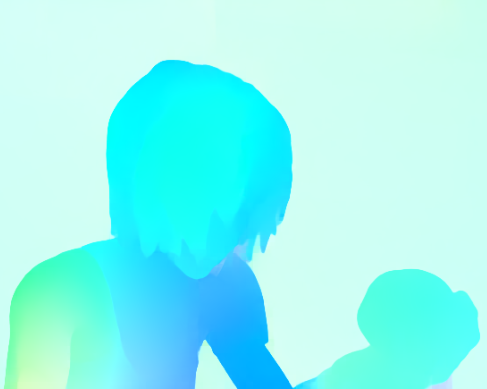} &
\includegraphics[width=0.235\linewidth]{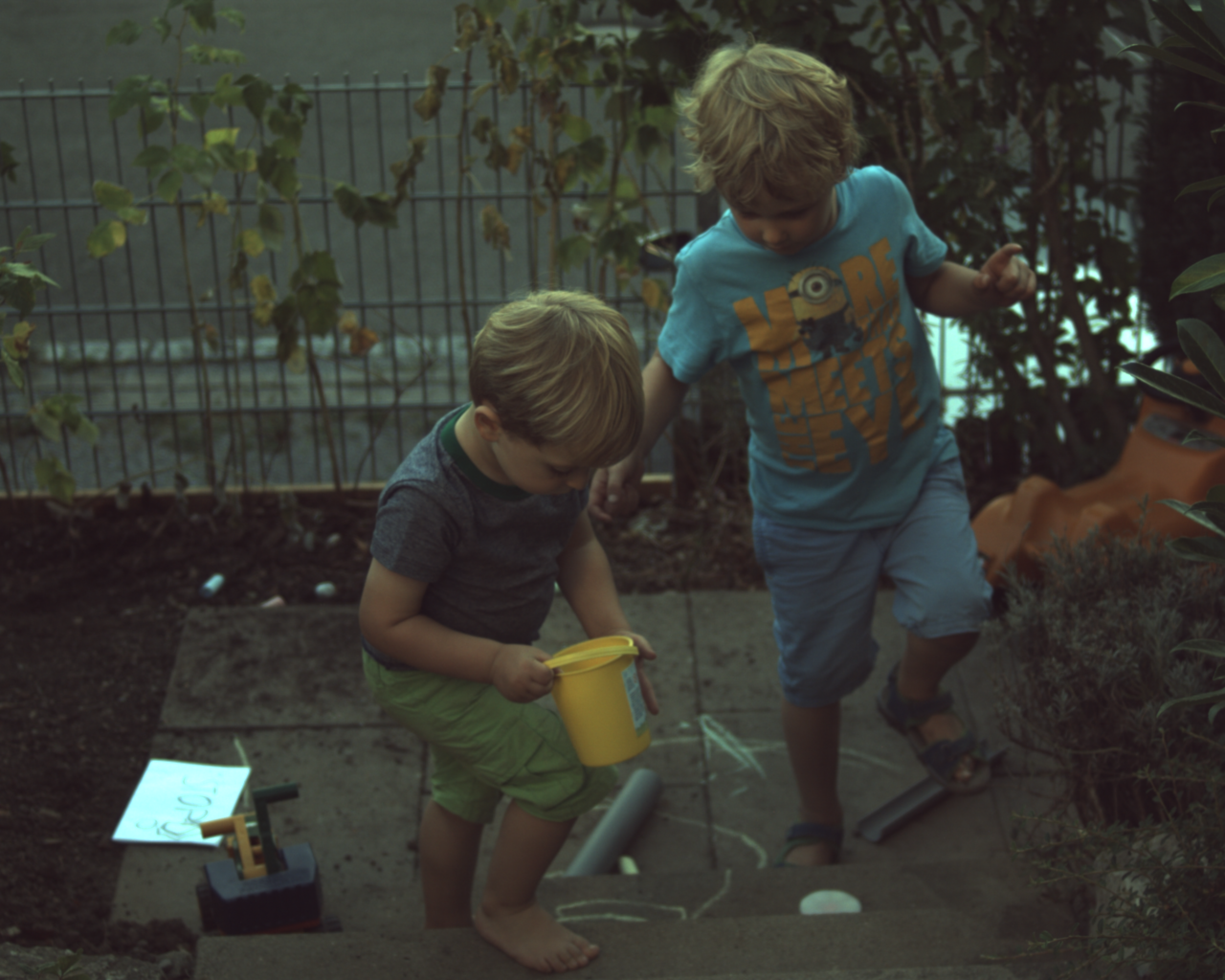} &
\includegraphics[width=0.235\linewidth]{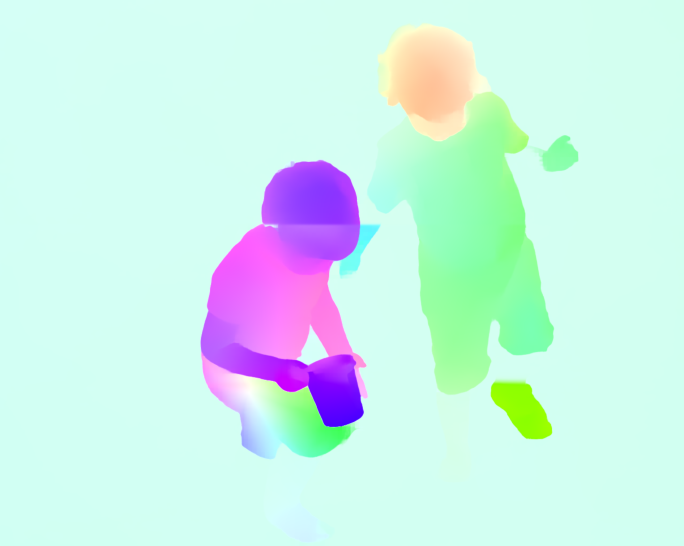} 
\end{tabular}
\caption{\textbf{How well does optical flow quality on current benchmarks generalise to real world optical flow quality?} Optical flow generalisability is, amongst others, measured on synthetic benchmarks \cite{huang2022flowformer, shi2023flowformer++, wang2024sea} such as Sintel \cite{butler2012naturalistic}. In this paper, we investigate if current-day benchmarks like Sintel can accurately measure generalisability to real-world settings. Both optical flow predictions above were made by the \textit{things} checkpoint of  FlowFormerPlusPlus \cite{shi2023flowformer++}. Whereas the prediction on the synthetic frame is nearly perfect, the real-world example reveals noticeable errors, particularly around the legs and parts of the arms, and a loss of fine detail in the hair. This leads us to ask the question: How well do current-day benchmarks measure generalisability to real-world benchmarks?} 
\label{fig:figuur-een}
\end{figure}

\section{Introduction}
Optical flow, the apparent 2D motion of pixels between consecutive frames, is a foundation for real-world applications ranging from action recognition and video editing to autonomous driving and robotics~\cite{alfarano2024estimating}. 
Modern optical flow methods~\cite{dosovitskiy2015flownet,sun2018pwc,teed2020raft,jiang2021learning,huang2022flowformer,shi2023flowformer++} learn from annotated data. Due to the difficulty of obtaining annotated per-pixel ground truth optical flow, the training data is created synthetically, such as in FlyingChairs~(C)~\cite{dosovitskiy2015flownet} and FlyingThings3D~(T)~\cite{mayer2016large} and their combination (C+T). For evaluating the generalisation of optical flow it is shown that synthetic pre-training can already yield competitive results \cite{huang2022flowformer} on differently distributed datasets such as Sintel \cite{butler2012naturalistic} and KITTI \cite{menze2015object}. A notable gap in evaluating optical flow generalisability is diverse real-world data, as illustrated in~\cref{fig:figuur-een}. Such real-world data is not used due to hardships of obtaining optical flow ground truth where obtaining pixel-accurate motion for every frame is not trivial, due to occlusions, non-rigid motion, lighting changes \etc. Manually annotating every pixel is essentially infeasible: even sparse tracking requires multiple annotator hours for only tens of points in a few seconds of video \cite{doersch2022tap}. Dense ground truth for real scenes therefore typically arises either from \emph{synthetic rendering} or from specialised hardware and reconstruction pipelines \cite{menze2015object, baker2011database, janai2017slow}.

\begin{figure}[ht]
    \centering
    \begin{minipage}{\linewidth}
        \centering
        \subcaption{Multiple confounders in existing benchmarks}
        \includegraphics[width=\linewidth]{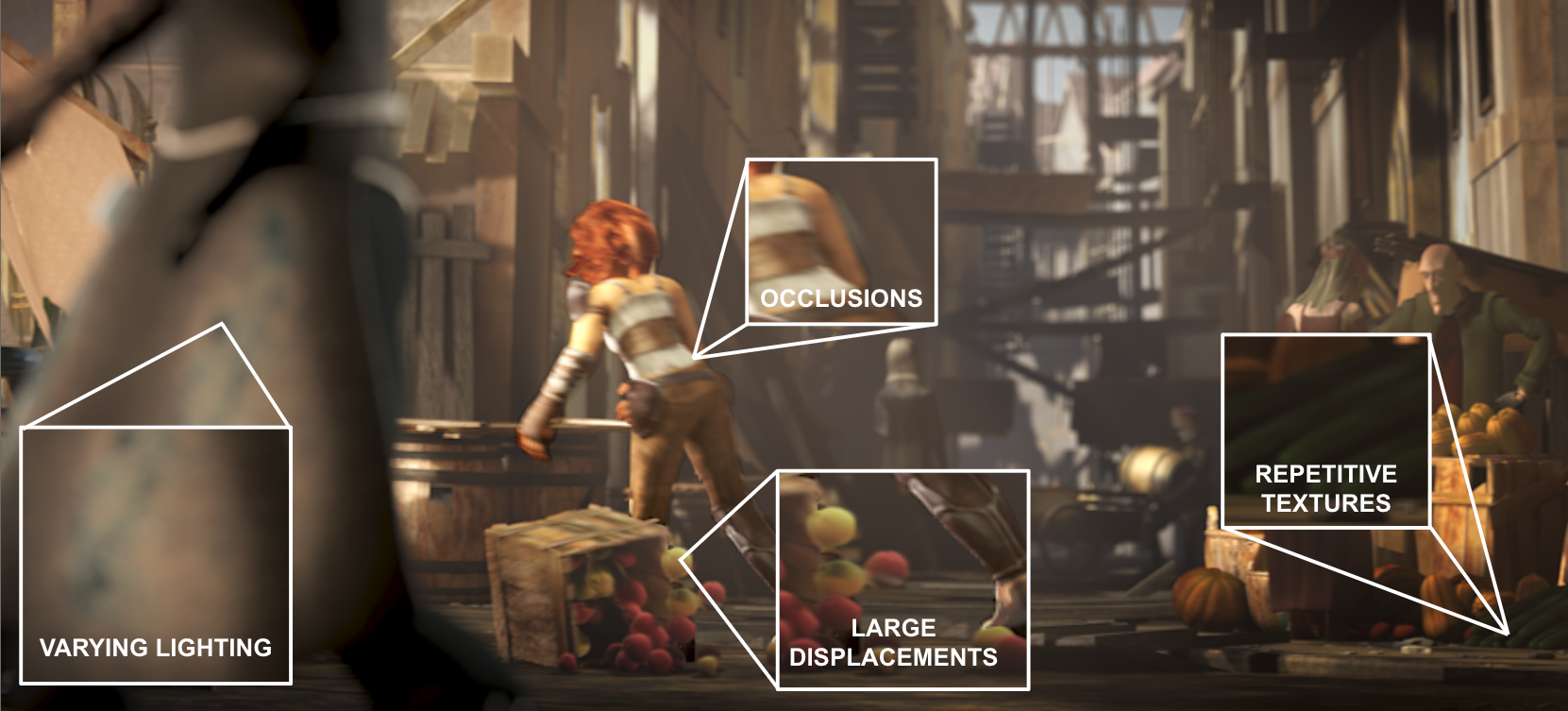}
    \end{minipage}

    \vspace{0.5em} 
    {\Large $\downarrow$} 

    \begin{minipage}{\linewidth}
        \centering
        \begin{minipage}{0.23\linewidth}
            \centering
            \subcaption*{Large Disps.}
            \includegraphics[width=\linewidth]{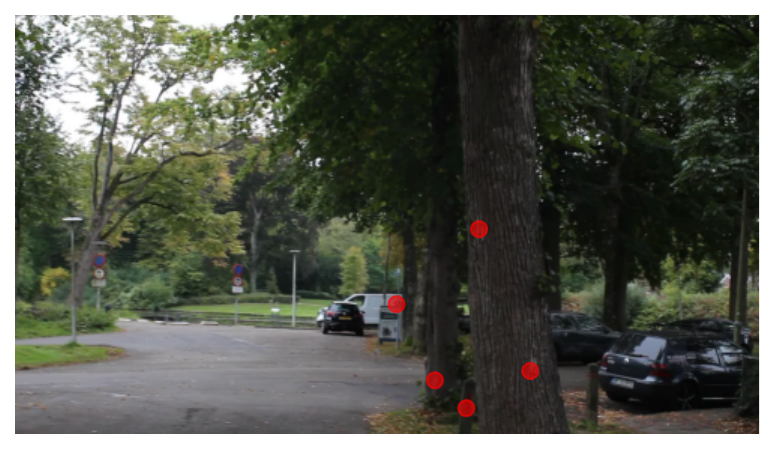}
        \end{minipage}
        \hfill
        \begin{minipage}{0.23\linewidth}
            \centering
            \subcaption*{Lighting}
            \includegraphics[width=\linewidth]{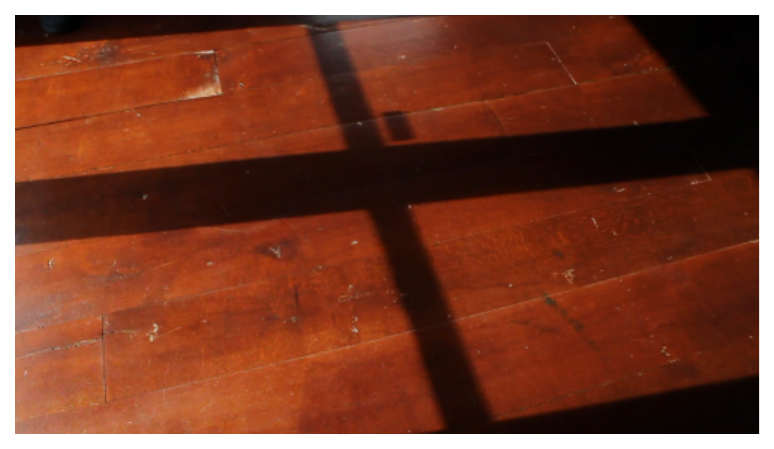}
        \end{minipage}
        \hfill
        \begin{minipage}{0.23\linewidth}
            \centering
            \subcaption*{Occlusions}
            \includegraphics[width=\linewidth]{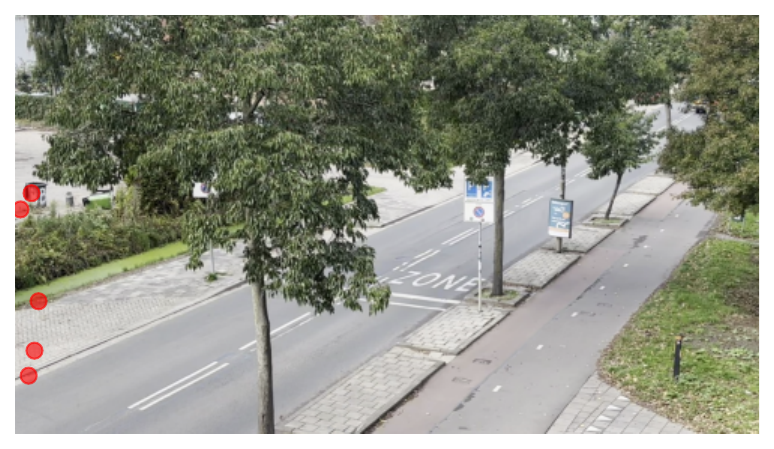}
        \end{minipage}
        \hfill
        \begin{minipage}{0.23\linewidth}
            \centering
            \subcaption*{Rep. Textures}
            \includegraphics[width=\linewidth]{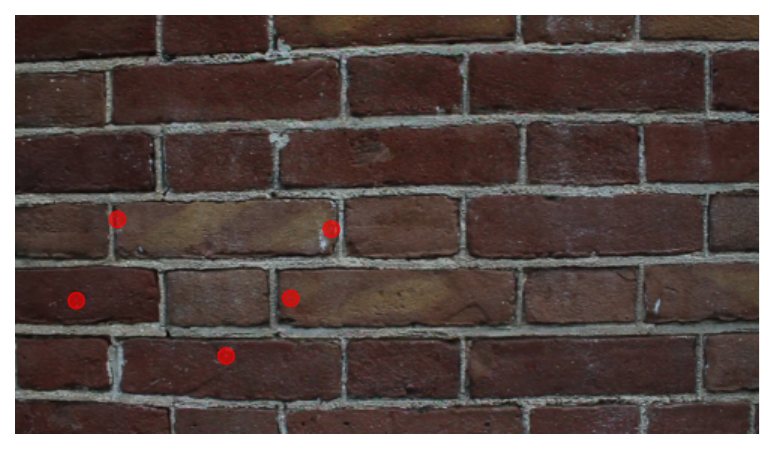}
        \end{minipage}
        
        \subcaption{FlowFactor controls for confounders}
    \end{minipage}

    \vspace{1em} 
    \caption{\textbf{Factorisation of Optical Flow failure modes.} (a) In existing optical flow benchmarks, multiple confounders, such as large displacements, occlusions, repetitive textures, and lighting variations, co-occur throughout the scene. This makes it difficult to attribute accuracy drops to specific confounders. (b) We instead isolate these confounders in a controlled real-world setting, enabling a more targeted analysis of known hard individual factors, that affect optical-flow accuracy, namely large displacements, repetitive textures, varying lighting, and occlusions. This dataset can then be used to pinpoint more fine-grained weaknesses of optical flow models.}
    \label{fig:figuur_twee}
\end{figure}

In this paper we ask the question whether or not optical flow models generalise to out-of-distribution (OOD) real-world data. This is important because in the real-world, it is not possible to finetune on in-distribution data, because of the difficulty of obtaining real-world optical flow annotations. Thus, OOD generalisation is often benchmarked by training on C+T and evaluating on more realistic datasets \cite{mehl2023spring, menze2015object, butler2012naturalistic}. These datasets are, however, narrow in domain, mainly driving cars, or are synthetic in nature. Recent literature \cite{agnihotri2025flowbench} suggests the generalisation ability of optical flow models has stagnated when it comes to synthetic corruptions. Here, we thus create our own benchmark, using broad, real-world data to estimate the generalisation gap. We incorporate in our real-world benchmark, three real-world datasets, namely Slow Flow \cite{janai2017slow}, TAP-Vid-DAVIS \cite{doersch2022tap}, and FlowFactor. We created the FlowFactor dataset to give a more fine-grained error metric, to improve understanding of optical flow failure modes. Most benchmarks report on broad metrics such as average end-point-error (EPE), where averaging over multiple confounders makes confounder-specific gains hard to measure. This makes model design harder, because it’s unclear what the model can actually improve. We create a broad real-world dataset to isolate different confounding factors, see~\cref{fig:figuur_twee}. We choose the following confounders, according to classical analyses and more recent overviews: occlusions, large displacements, illumination changes, and complex textures. As they remain enduring challenges for optical flow models \cite{zheng2020optical, eslami2024rethinking, zhao2022global, huang2022flowformer, shi2023flowformer++, zhai2021optical}. This leads to our FlowFactor dataset. Our main contributions are:
\begin{itemize}
    \item We introduce \textit{FlowFactor}, a new sparsely annotated real-world optical flow benchmark designed around four isolated challenge factors: occlusions, lighting changes, repetitive textures, and large displacements. This enables factor-specific analysis of failure modes rather than a single aggregate error.
    \item Using FlowFactor, we provide a detailed analysis of where current models fail and highlight directions for developing optical flow methods that generalise more reliably to real-world applications.    
     \item We create a real-world optical flow evaluation benchmark containing 8,204 samples. We transform the TAP-Vid-DAVIS \cite{doersch2022tap} dataset into an optical flow dataset, and combine this with the Slow Flow \cite{janai2017slow} and FlowFactor datasets into a benchmark. Using this benchmark we do a comprehensive benchmark of the current state of real-world optical flow accuracy.

\end{itemize}

\section{Related Work}
\label{sec:related}

\subsection{Optical Flow Datasets and Benchmarks}
\label{sec:rw_datasets}
Early evaluation efforts date back to Barron \etal \cite{barron1994performance}, combining both synthetic data from the Yosemite sequence \cite{heeger1987model} and simple real toy datasets. Later, the Middlebury benchmark emerged as a relatively small, laboratory-style dataset \cite{baker2011database}. While crucial for methodology development, these datasets are limited in resolution, diversity, and motion complexity.

The large-scale data requirements of deep learning triggered a wave of synthetic datasets. FlyingChairs \cite{dosovitskiy2015flownet} renders chairs undergoing planar motions over random natural-image backgrounds to support FlowNet training \cite{dosovitskiy2015flownet}. Mayer \etal \cite{mayer2016large} later introduced a variety of optical flow datasets, including FlyingThings3D, which simulates everyday objects moving in 3D. These datasets remain the standard for pre-training, often referred to as C+T (FlyingChairs + FlyingThings3D) in recent works. Customisable dataset generators such as Kubric \cite{greff2022kubric} allow users to synthesise many variants of scenes and tasks, including optical flow, by randomising scene content and rendering parameters. Autoflow~\cite{sun2021autoflow} allows for learning a dataset, based on the accuracy on a target dataset. This dataset is learned by a hyperparameter search using evolutionary algorithms. All of these pre-training datasets are synthetic in nature. This leads to a generalisation gap to the real-world. 

Evaluation benchmarks tend to be more realistic in nature. MPI Sintel \cite{butler2012naturalistic} is based on an open-source animated film and includes challenging effects such as motion blur, long-range motions, and non-rigid deformations. KITTI 2012/2015 \cite{geiger2012we,menze2015object} provide real driving scenes, where optical flow ground truth is obtained semi-automatically by using LIDAR and camera parameters to estimate non-moving objects, and fitting 3D CAD models to moving vehicles. HD1K \cite{kondermann2016hci} extends real-world driving benchmarks to a higher resolution and more diverse weather and illumination. More recently, Spring \cite{mehl2023spring} offers a large, high-resolution, computer-generated benchmark for stereo, optical flow, and scene flow, drawn from a photorealistic Blender short film with 4$\times$ super-resolved ground truth. Slow Flow \cite{janai2017slow} is a real-world benchmark that takes a hardware-driven approach by using high-speed cameras. By exploiting the approximate linearity of small motions, the authors reconstruct accurate reference flow fields for a variety of real-world scenes. 

Other specialised datasets focus on particular phenomena such as low-light conditions with varying brightness and noise \cite{zheng2020optical} or robustness to a broad range of visual corruptions \cite{schmalfuss2025robustspring}. 

While this ecosystem of benchmarks is large and diverse, most of them report aggregate metrics, such as EPE, or only coarse breakdowns such as occluded vs.\ non-occluded regions, foreground vs.\ background, or broad displacement bins. More fine-grained statistics, such as rigid vs. non-rigid regions and low vs. high texture, are beginning to appear \cite{mehl2023spring}, but they are not measured in isolation: multiple confounding factors change simultaneously, making it difficult to attribute failures to specific causes. Our work is complementary in that FlowFactor is explicitly designed to isolate individual factors: occlusions, lighting changes, repetitive textures and large displacements; in a controlled, real-world setting.

\subsection{Sparse Motion Annotations and Point Tracking}
Obtaining dense ground truth for arbitrary real videos remains prohibitively expensive, motivating research on sparse annotations. Over the years, various real-world optical flow datasets have used human involvement to different degrees: fitting CAD models to moving objects \cite{menze2015object}, manually selecting object contours and interpolating flow inside \cite{liu2008human,zhou2024adverse,li2021gyroflow}, annotating motion boundaries \cite{weinzaepfel2015learning}, or directly annotating individual pixels \cite{doersch2022tap}. These approaches exploit the human visual system’s sensitivity to motion, depth ordering, and object grouping \cite{sturzel2004perceptual}. Our FlowFactor dataset, the TAP-Flow dataset, and the densely labeled Slow Flow benchmark~\cite{janai2017slow} combined, yield a suite of real-world datasets that cover both sparse and dense labels across diverse conditions. For our FlowFactor dataset we choose sparse manual labeling because it is the most versatile approach and supports annotation of non-rigid motion.

\subsection{Robustness, generalisation, and factorised evaluation}
Due to the use of synthetic training data, several surveys and benchmarks have begun to question the robustness and generalisation of optical flow models. Specialised benchmarks probe particular phenomena, such as low-light conditions \cite{zheng2020optical} or robustness to synthetic corruptions including noise, blur, and compression \cite{schmalfuss2025robustspring}. FlowBench \cite{agnihotri2025flowbench} explicitly aims to benchmark in-distribution accuracy, reliability, and generalisation via corruption and adversarial attacks, and reports that as accuracy on standard benchmarks has increased, reliability and robustness have not necessarily followed.

Another line of work focuses on generalisation across datasets. Many optical flow models claim strong distribution generalisation by training on less realistic synthetic datasets \cite{ilg2017flownet,mayer2016large} and evaluating on more realistic validation sets such as Sintel \cite{butler2012naturalistic}, Spring \cite{mehl2023spring}, or KITTI~\cite{menze2015object}. The Robust Vision Challenge \cite{robustvisionchallenge} and related efforts encourage evaluation on multiple benchmarks to assess cross-dataset robustness reporting on the mean error on all of these datasets combined, though hyperparameter tuning is allowed on those benchmarks beforehand, making it difficult to fully disentangle generalisation from dataset-specific tuning \cite{bauergeneralization}.

Closest to our work are efforts that analyse cross-dataset generalisation of modern deep optical flow models \cite{teed2020raft,xu2022gmflow,agnihotri2025flowbench}. However, these datasets are often narrow in domain \cite{butler2012naturalistic, kondermann2016hci} or synthetic in nature \cite{butler2012naturalistic}. Moreover, these evaluations typically aggregate accuracy over entire datasets and do not systematically isolate individual failure factors in real scenes. Also, they often rely on a single real-world benchmark (e.g., KITTI \cite{menze2015object} or Slow Flow \cite{janai2017slow}) and use test data that is closely related to the training data \cite{agnihotri2025flowbench}. In contrast, our work focuses on real-world evaluation suites that are unseen during training and are explicitly designed (in the case of FlowFactor) to factorise evaluation into four isolated challenges: occlusions, lighting changes, repetitive textures, and large displacements. This allows us to move beyond aggregate error measures and towards a more diagnostic understanding of the root failure causes of current optical flow models.

\begin{figure}[t]
\centering
\begin{subfigure}{0.48\linewidth}
    \centering
    \includegraphics[width=0.49\linewidth]{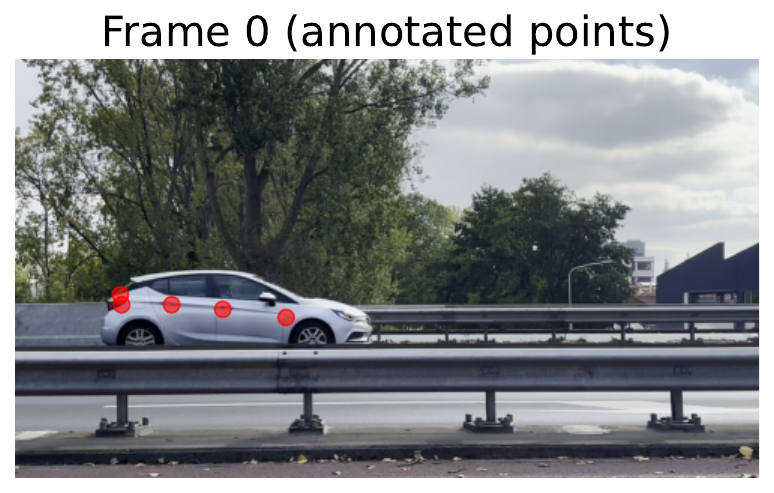}
    \includegraphics[width=0.49\linewidth]{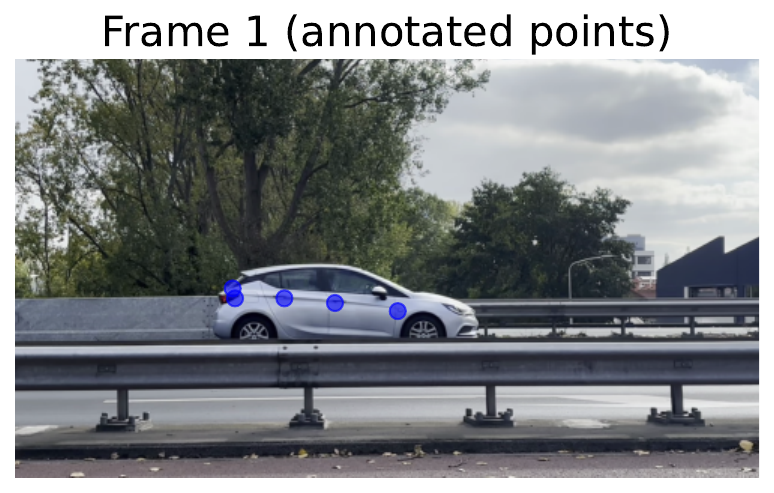} \\
    \includegraphics[width=0.49\linewidth]{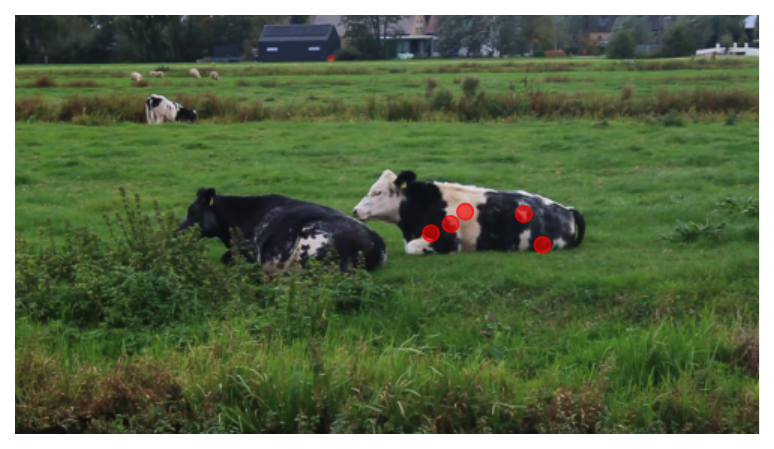}
    \includegraphics[width=0.49\linewidth]{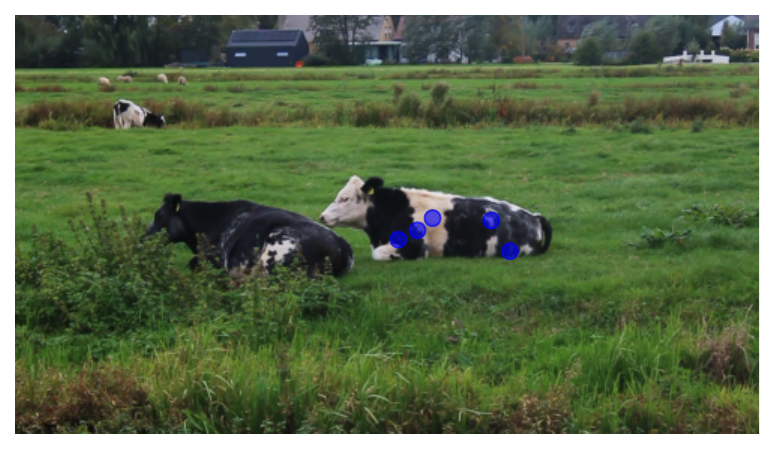}
    \caption{\textbf{Large displacement}. Objects move more than 25 pixels, exceeding the motion magnitude typically observed in common benchmarks \cite{butler2012naturalistic, menze2015object, mehl2023spring}.}
\end{subfigure}
\hfill
\begin{subfigure}{0.48\linewidth}
    \centering
    \includegraphics[width=0.49\linewidth]{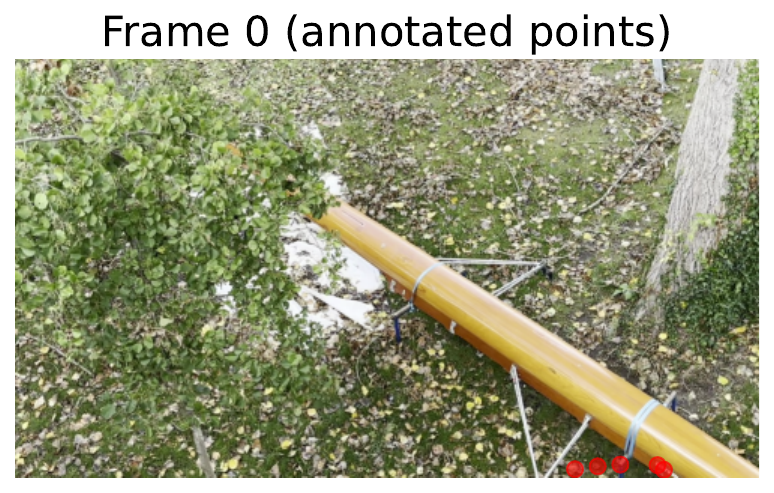}
    \includegraphics[width=0.49\linewidth]{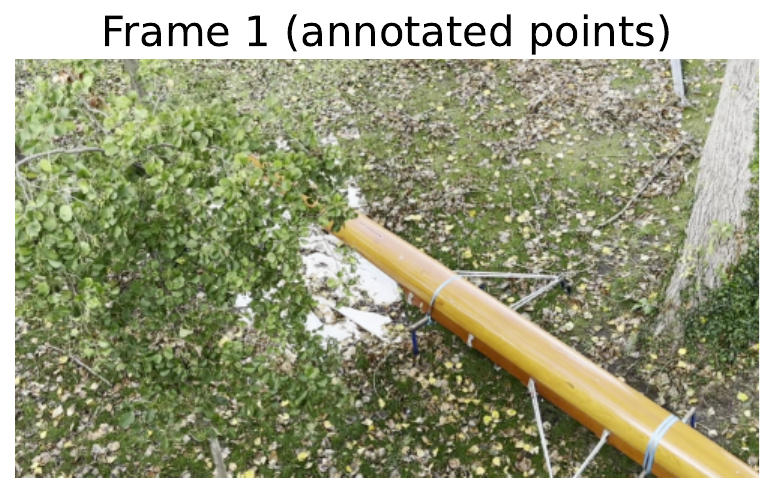} \\
    \includegraphics[width=0.49\linewidth]{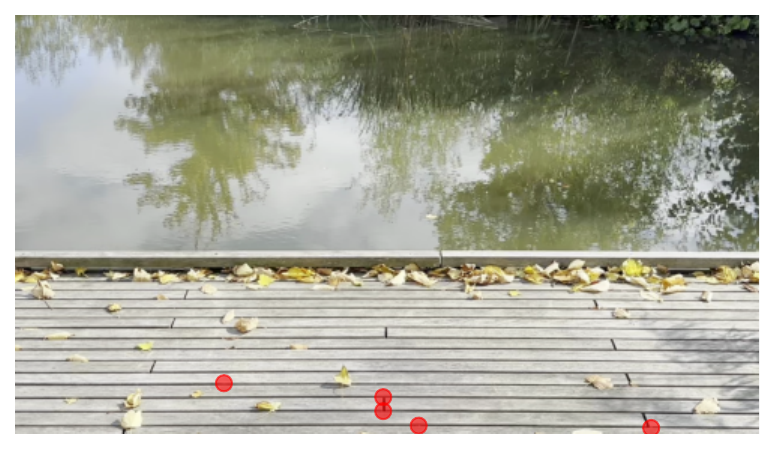}
    \includegraphics[width=0.49\linewidth]{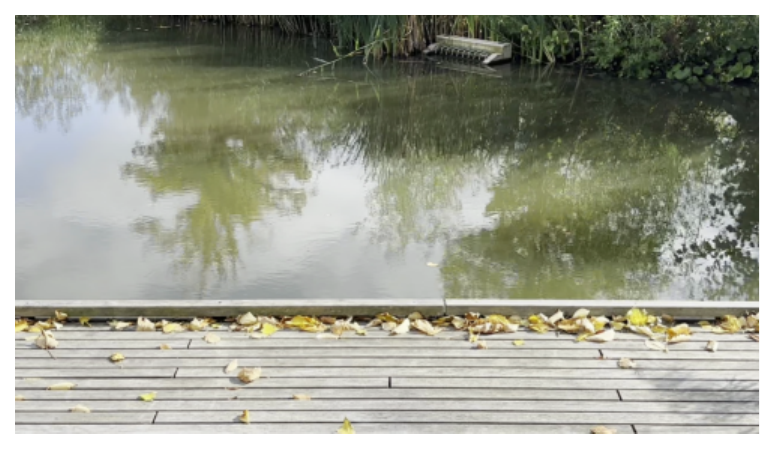}
    \caption{\textbf{Occlusion}. Annotated pixels move outside the frame in the second image and correspondences must then be inferred from global context.}
\end{subfigure}
\begin{subfigure}{0.48\linewidth}
    \centering
    \includegraphics[width=0.49\linewidth]{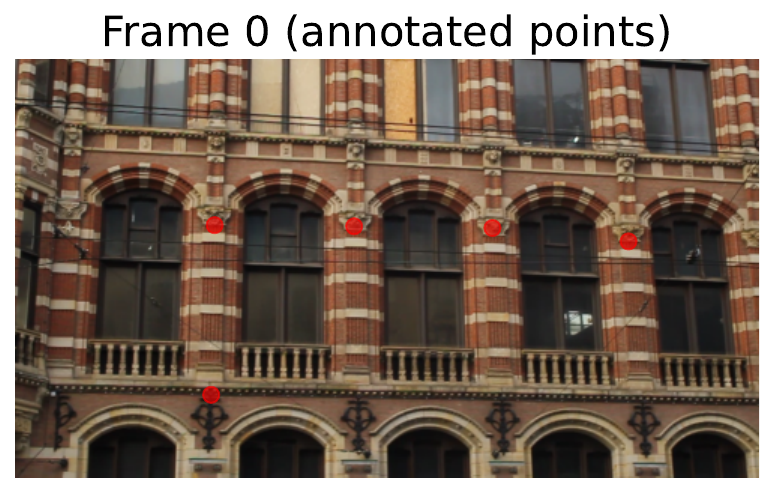}
    \includegraphics[width=0.49\linewidth]{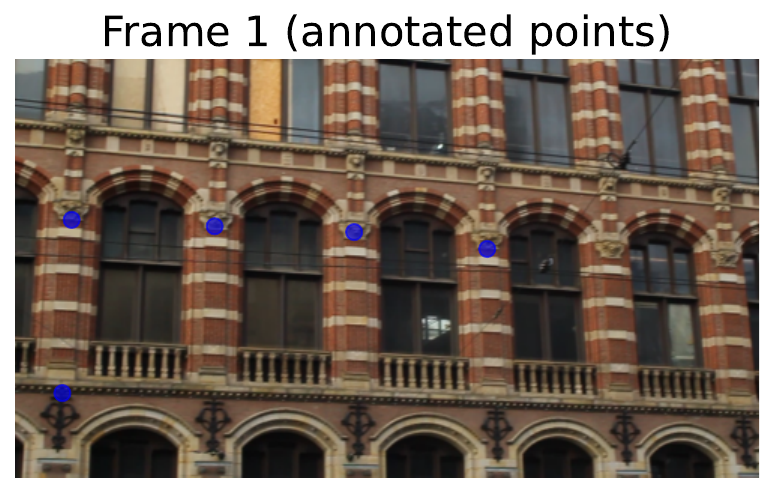} \\
    \includegraphics[width=0.49\linewidth]{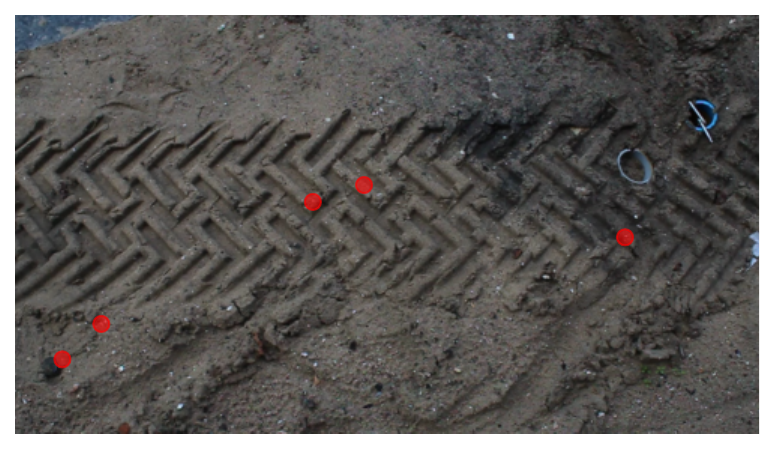}
    \includegraphics[width=0.49\linewidth]{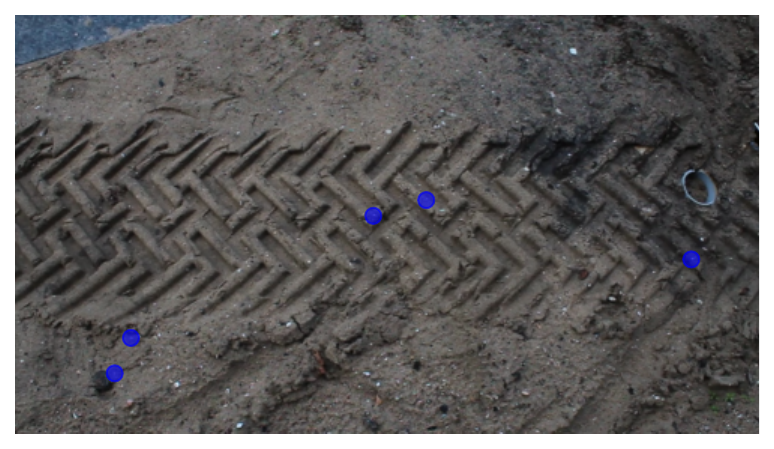}
    \caption{\textbf{Repetitive texture}. Many spurious correlations exist in this scene, aiming to test the model's ability to differentiate between fine-grained differences in a scene and use global context.}
\end{subfigure}
\hfill
\begin{subfigure}{0.48\linewidth}
    \centering
    \includegraphics[width=0.49\linewidth]{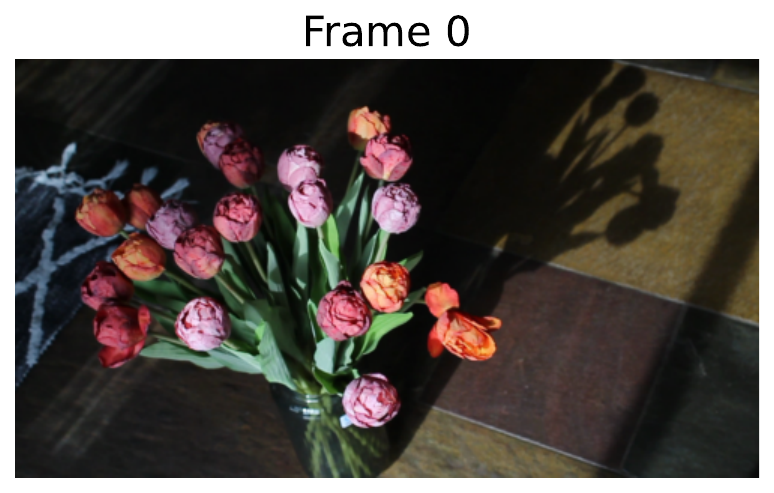}
    \includegraphics[width=0.49\linewidth]{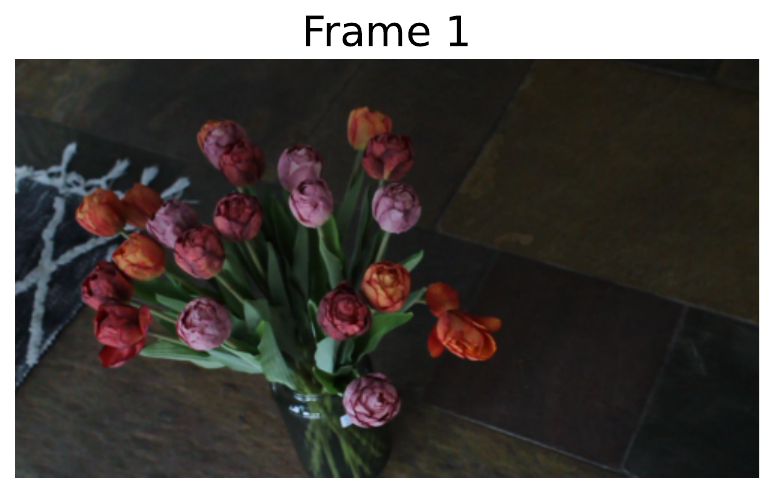} \\
    \includegraphics[width=0.49\linewidth]{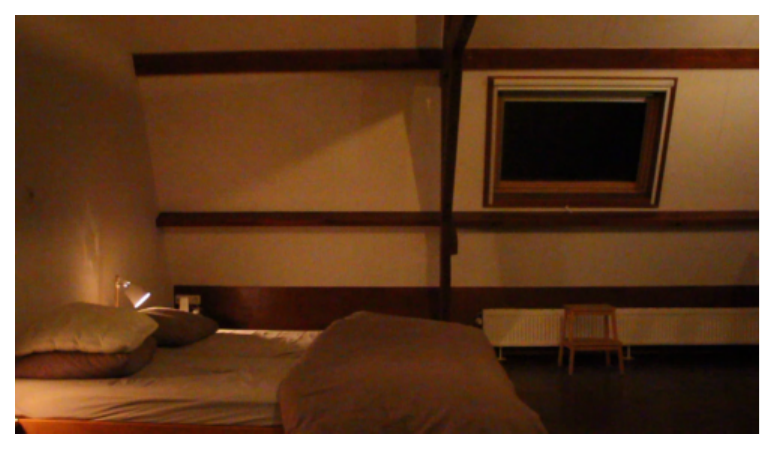}
    \includegraphics[width=0.49\linewidth]{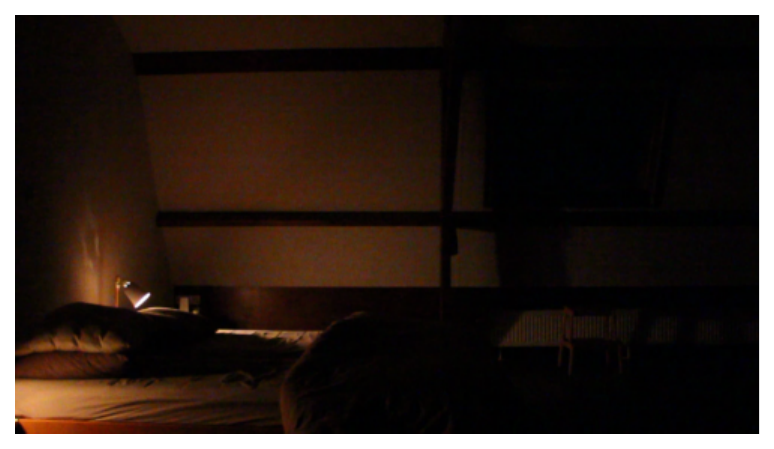}
    \caption{\textbf{Lighting changes}. All pixels are valid, the flowfield is completely zero. The appearance changes but the shape remains completely the same. Thereby benchmarking the model's ability to interpret geometrical content of a scene.}
\end{subfigure}

\caption{Example frame-pairs from the FlowFactor dataset. Annotated points are best viewed zoomed in.}
\label{fig:flowfactor_examples}

\end{figure}

\section{Datasets for real-world generalisation}
We construct three datasets to benchmark how well optical flow models generalise to the real-world. The \textit{FlowFactor} dataset allows for precise controlled settings, while \textit{TAP-flow} \cite{doersch2022tap} captures an uncontrolled setting, with multiple challenging confounders, Slow Flow \cite{janai2017slow} allows for a dataset that also incorporates motion blur and has a dense flow field. 

\subsection{FlowFactor: Isolated Category Benchmark}
To gain insight into failure cases we sparsely annotated 1,000 frame pairs at a resolution of 1280×720, grouped into four distinct categories: large displacements, repetitive textures, occlusions, and lighting variation. These factors are selected based on common failure modes in optical flow \cite{baker2011database} and target different challenges, which can be encountered in real-world use. The isolated settings allow for controlled model testing. Examples can be seen in~\cref{fig:flowfactor_examples}.

\paragraph{Large displacements.} Large displacements evaluate whether a model can accurately track substantial motion. We select cases where the Euclidean distances between pixel pairs are $\geq 25$ pixels, but typically is much larger. This displacement can be achieved by object motion, camera motion, or both object motion and camera motion.

\paragraph{Occlusions.} Occlusions assess whether a model can reason about objects that become partially hidden, either by other objects or by the image boundary.
We include 125 pairs with out-of-frame occlusions, where the object leaves the image boundary, and 125 pairs with inter-object occlusions. For the inter-object case we take visually flat surfaces that get partially occluded (e.g. a tree in front of a car) and we linearly interpolate occluded points on the surface. 

\paragraph{Repetitive Textures.} Repetitive textures test the ability to resolve visual ambiguity when local appearance is insufficient. As the local region of a pixel remains visually similar, it requires the model to use global context to solve the task, as it cannot be solved by local context.
We select 250 frame pairs containing visually similar but spatially shifted textures. This produces ambiguous local appearance, requiring global context for correct correspondence.

\paragraph{Lighting Variation.}
Lighting variation tests robustness to changes in illumination that alter appearance while leaving the underlying motion unchanged. We select 250 static frame pairs where lighting intensity, shadows, or glare change across frames while all objects remain stationary. These pairs contain dense zero-flow fields but look visually different.

\subsubsection{Pixel selection strategy. } 
\begin{figure}[t]
    \centering
    \begin{subfigure}{0.4\linewidth}
        \centering
        \includegraphics[width=\linewidth]{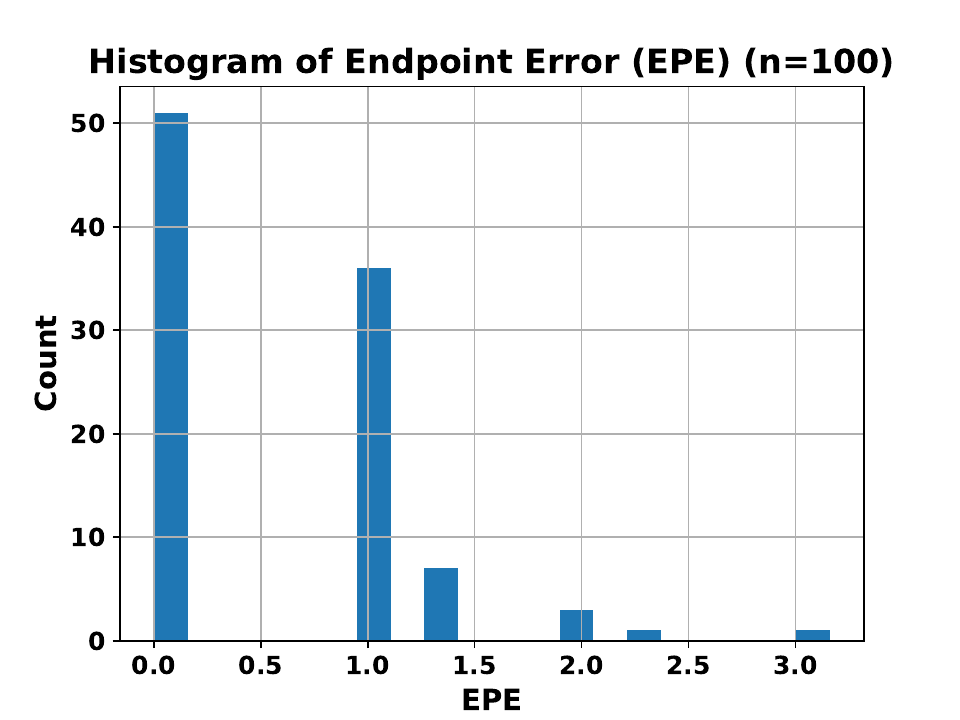}
        \caption{Annotation consistency.}
        \label{fig:pixel-exp}
    \end{subfigure}
    \hfill
    \begin{subfigure}{0.25\linewidth}
        \centering
        \includegraphics[width=\linewidth]{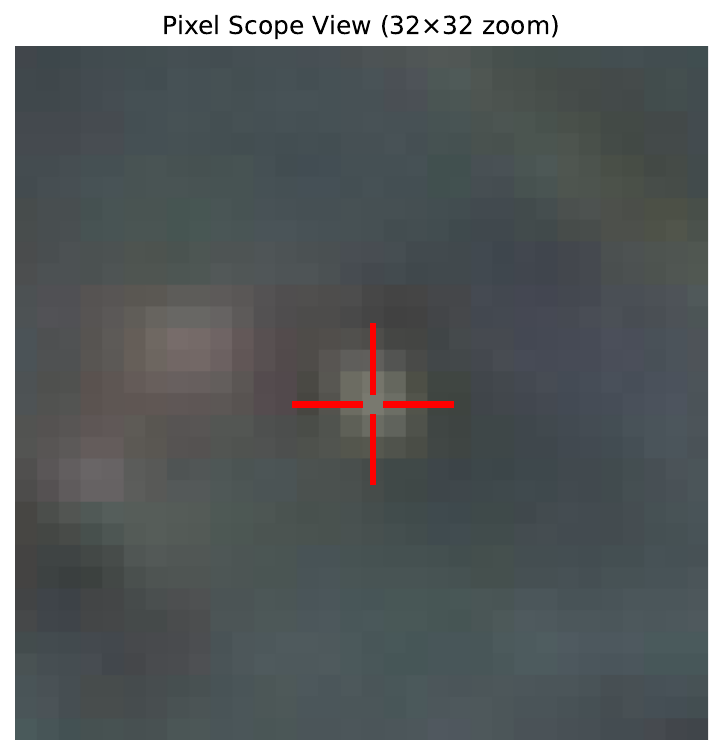}
        \caption{Zoomed example annotation of a bright spot}
        \label{fig:disp10}
    \end{subfigure}
    \hfill
    \begin{subfigure}{0.25\linewidth}
        \centering
        \includegraphics[width=\linewidth]{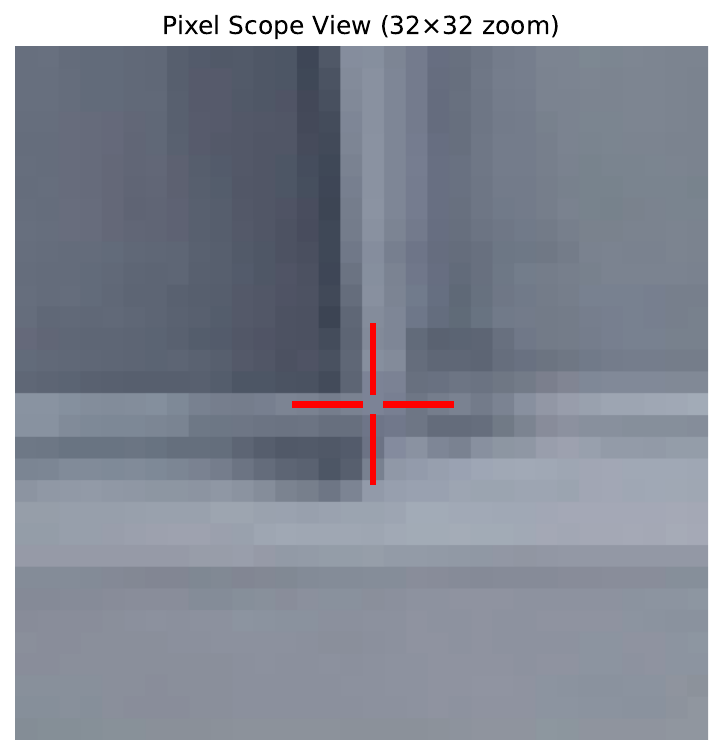}
        \caption{Zoomed example annotation of an edge}
        \label{fig:disp11}
    \end{subfigure}

    \caption{ \textbf{Feasibility and accuracy of hand-annotated points.} (a) Histogram of Euclidean distances, when 100 of our points are annotated by others. The average distance is 0.58, which falls within our used 3-pixel error metric. (b and c) We annotate easily recognisable pixels like edges and bright spots. }
    \label{fig:combined-pixel-analysis}
\end{figure}

We follow the annotation strategy of Doersch \etal \cite{doersch2022tap} by manually selecting five informative points per pair. Annotators are free to choose points of interest, resulting in natural diversity in difficulty, location, and appearance. The points that are chosen are easily annotatable such as bright spots and edges, examples can be seen in~\cref{fig:combined-pixel-analysis}. We show the correctness of our annotation by asking five people to annotate up to 20 points per person resulting in an average 0.58 EPE to our chosen points, way below the three point threshold, which is the margin in our Fl-error metric. The results can be seen in~\cref{fig:pixel-exp}.

A custom annotation interface exports the results in KITTI-style flow maps \cite{Ge2025, Petre2025, Timmerije2025, Klijnsma2025, Dahal2025}. We use 5 points per scene as this gives a strong indication of the accuracy of a model on the entirety of the scene relative to other models. 

We verify this hypothesis by using the first two frame pairs of each TAP-Vid-DAVIS \cite{doersch2022tap} video and sampling 2 sets of 5 non-overlapping annotated points from the same frame. This leads to 2 datasets with 60 frame pairs each. Each dataset uses the same scenes but different points. We hypothesise that the relative ordering of the models remain roughly the same between both datasets, thereby showing that 5 points can be used as a strong proxy for the relative accuracy on the entire scene. This appears to be true as the spearman correlation coefficient is 0.95 (p = 3.51e-22), which suggests a strongly positive correlation between the ranking of both point sets. Thereby providing substance to the claim that 5 points are a good proxy of relative scene accuracy.

\subsection{TAP-Flow: Uncontrolled benchmark  }
We depart from the high-quality TAP-Vid-DAVIS point-tracking benchmark \cite{doersch2022tap}. This dataset contains 30 videos with up to 5 objects with 5 point annotations per object labelled. 
\paragraph{Optical Flow Repurposing.}
The original TAP-Vid dataset \cite{doersch2022tap} includes varying lighting, large displacements, varied scenery and a diverse flow field, and thus is well-suited for evaluating real-world optical flow generalisation. Because TAP-Vid consists of sparse point tracks rather than flow fields, we repurpose it for optical flow evaluation, we use all the 30 TAP-Vid-DAVIS \cite{doersch2022tap} videos and pick the points that appear in both frames, so no occluded points are used, and use an interval between frames ranging from 3 to 6 for every frame. Resulting in 3757 frames across 30 videos. This benchmark allows for a balanced marker of real-world accuracy. These sparse annotations are exported in KITTI-style format, enabling standardised optical flow evaluation.

\subsection{Slow Flow: uncontrolled benchmark} We also introduce the Slow Flow dataset~\cite{janai2017slow}. Slow Flow, has 3 different categories, in which the 90th percentile displacement is 100, 200, and 300 pixels, with blur lengths of 0, 1, 3, 5, 9 frames. This results in 3447 dense ground-truth frames \cite{janai2017slow}. Thereby being complementary to the TAP-Flow \cite{doersch2022tap} dataset as TAP-Flow does not contain motion blur and occlusions, leading to a more diverse benchmark.
\section{Evaluation methodology}
\subsection{Model selection}
We use the models provided by the PTLFlow framework \cite{morimitsu2021ptlflow}. To establish correctness of the implementation we replicate reported benchmark accuracies on the training set of both KITTI \cite{menze2015object} and Sintel \cite{butler2012naturalistic}. We eliminate models which differ in our reproduction more than 2.0 epe on Sintel \cite{butler2012naturalistic} or 2.0 Fl-all on KITTI \cite{menze2015object}, than reported by their respective authors. Because of this reason we eliminated the HD3 model, \cite{yin2019hierarchical}, the RAFT\_small model \cite{teed2020raft}, and the MaskFlowNet\_small model \cite{zhao2020maskflownet}. The deviations between reported errors by their authors and our implementation can be found in the supplemental material.

\subsection{Checkpoint selection}
We use the Things (C+T) \cite{teed2020raft, zhao2022global} checkpoint for all models corresponding to a training regime of first training on FlyingChairs~\cite{dosovitskiy2015flownet} and then on FlyingThings3D~\cite{mayer2016large}. The checkpoints are provided by the authors of their respective papers and hence training conditions may have differed. Training all models is however infeasible and would take up substantial computational resources. We use the Things checkpoint because it's the checkpoint that nearly every model provides and is the one most used to measure OOD generalisability \cite{huang2022flowformer, shi2023flowformer++, wang2024sea, teed2020raft}. 

\subsection{Error metric}
We use the Fl-all metric propagated by the authors of KITTI \cite{menze2015object}. Fl-all counts a pixel as correct if the Euclidean distance is less than 3 pixels or 5\% to the ground truth pixel. The Fl-all score is computed by the PTLFlow framework \cite{morimitsu2021ptlflow}. The rationale for the Fl-all error is to account for human error.  This is based on the assumption that human error falls within a range of 3 pixels, see~\cref{fig:combined-pixel-analysis}. And that to have a more specific metric, such as the End-Point-Error which is the Euclidean distance, would suggest a level of detail that is not present in the ground truth. Since we also make use of human annotations, we apply the Fl-all metric consistently across all of our results.
\section{Experiments}

\subsection{Exp. 1: How well does optical flow quality on current benchmarks generalise to real world optical flow quality?}
\begin{figure}[t]
  \centering
  \includegraphics[width=0.33\columnwidth]{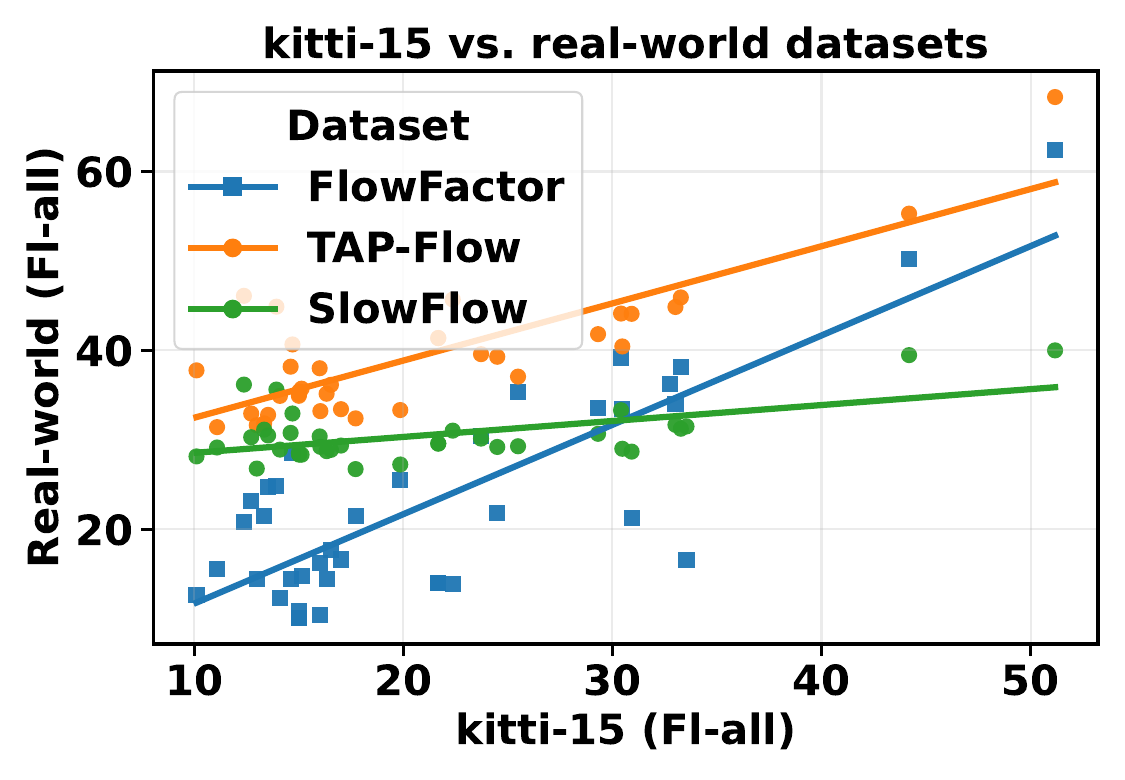}%
  \includegraphics[width=0.33\columnwidth]{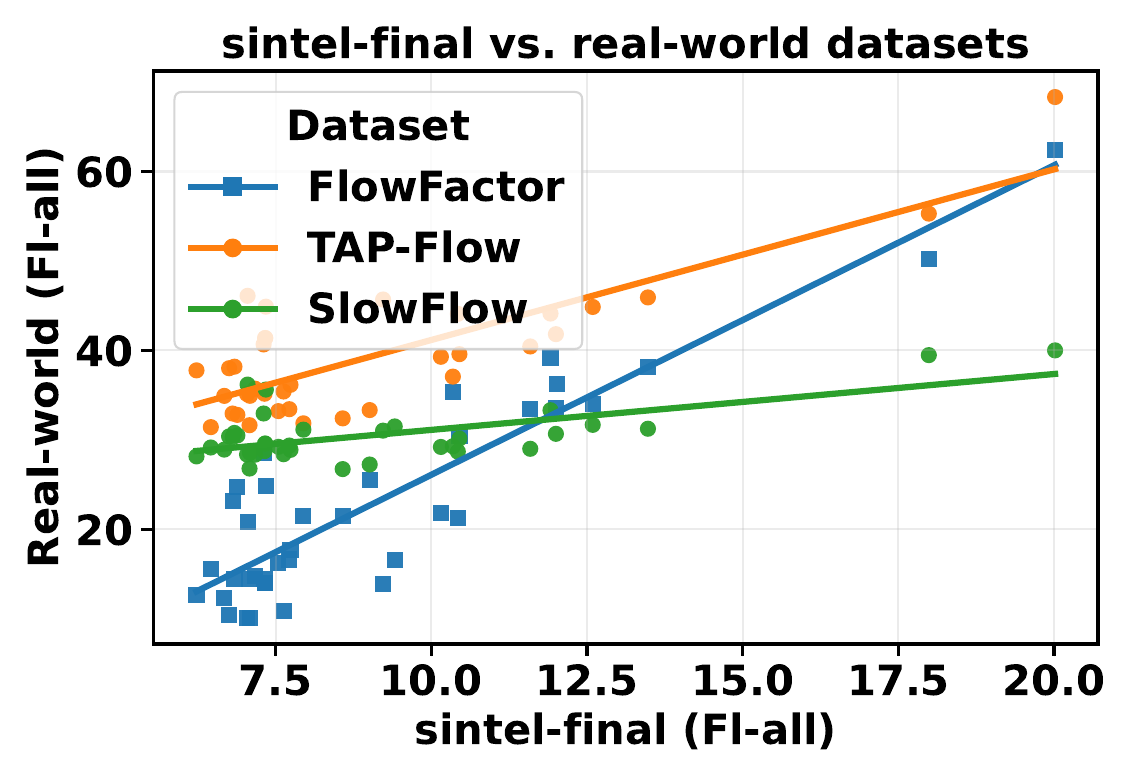}
  \includegraphics[width=0.33\columnwidth]{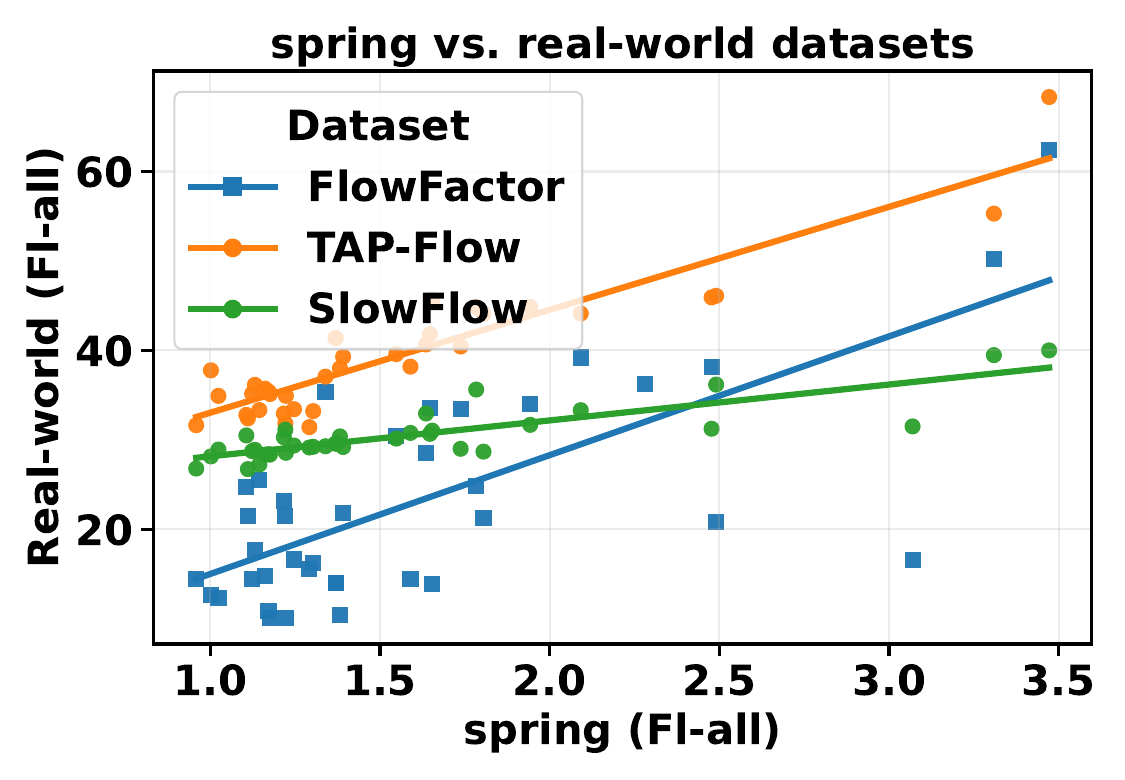}
  \caption{\textbf{Correlation between real-world datasets and KITTI \cite{menze2015object},  Sintel-Final \cite{butler2012naturalistic} and Spring \cite{mehl2023spring} (all models).}
  We plot the Fl-error of every Things model and the Ordinary Least Squares (OLS) fit on the KITTI \cite{menze2015object}, Sintel-Final \cite{butler2012naturalistic} and Spring \cite{mehl2023spring} training sets against three real-world benchmarks: FlowFactor, Slow Flow \cite{janai2017slow} and TAP-Flow dataset \cite{doersch2022tap}. The overall downward trend indicates that, when considering all models, these benchmarks still serve as a reasonable proxy for estimating real-world applicability.}
  \label{fig:generalisability_OLS_all}
\end{figure}

We first evaluate all models on three real-world benchmarks: FlowFactor, Slow Flow \cite{janai2017slow} and TAP-Flow \cite{doersch2022tap} to obtain a balanced estimate of real-world accuracy, the results are shown in Figure \ref{fig:generalisability_OLS_all}. When considering all models, the overall trend is downward on all benchmarks, indicating that a lower error on KITTI \cite{menze2015object}, Sintel \cite{butler2012naturalistic} and Spring \cite{mehl2023spring} generally results in a lower error on real-world datasets.

\begin{figure}[t]
  \centering
  \includegraphics[width=0.33\columnwidth]{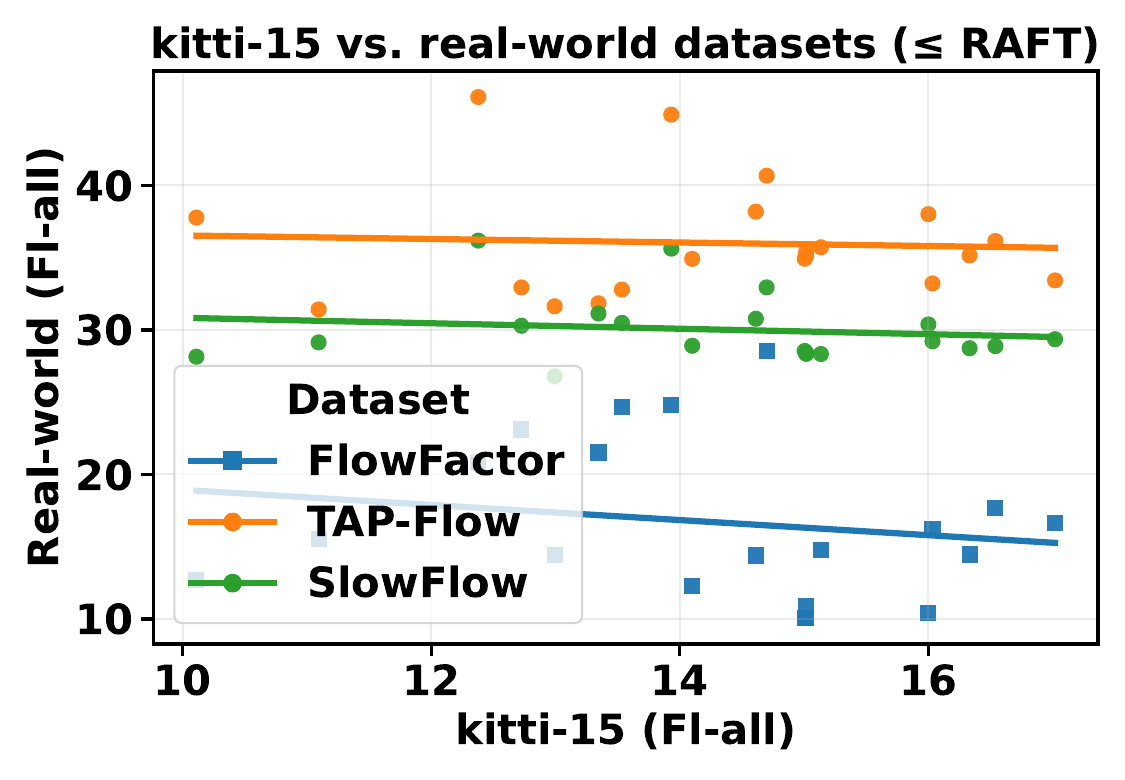}%
  \includegraphics[width=0.33\columnwidth]{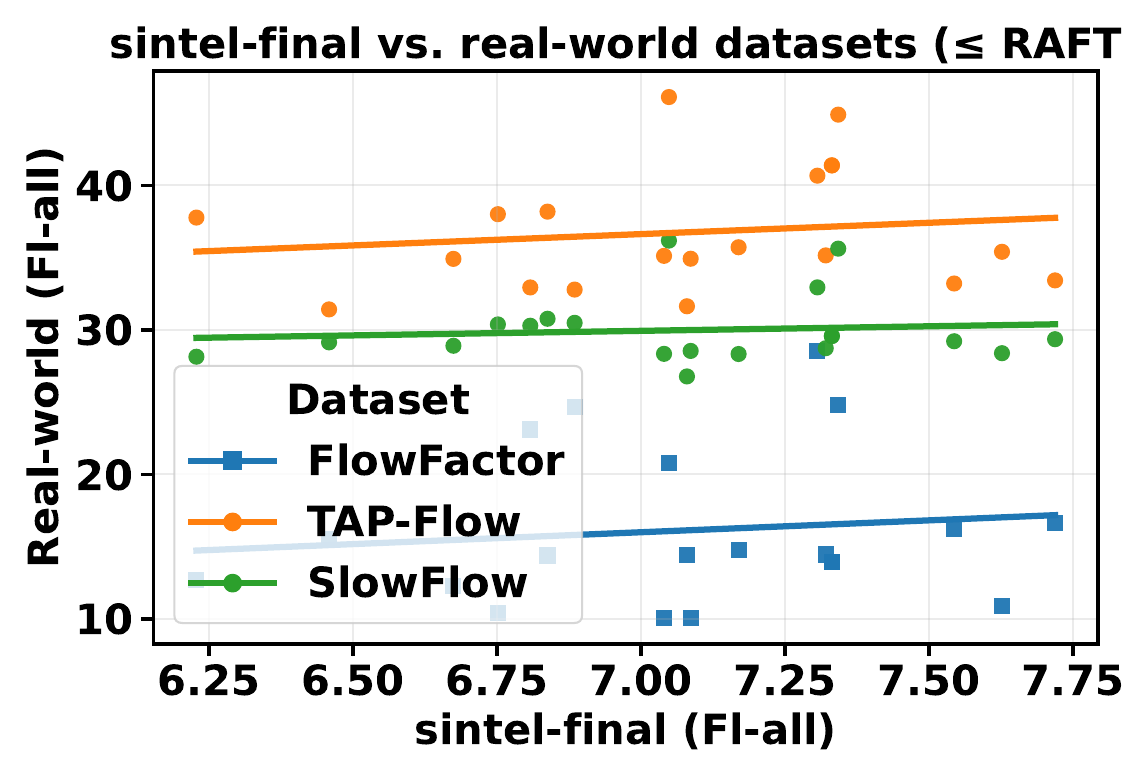}
  \includegraphics[width=0.33\columnwidth]{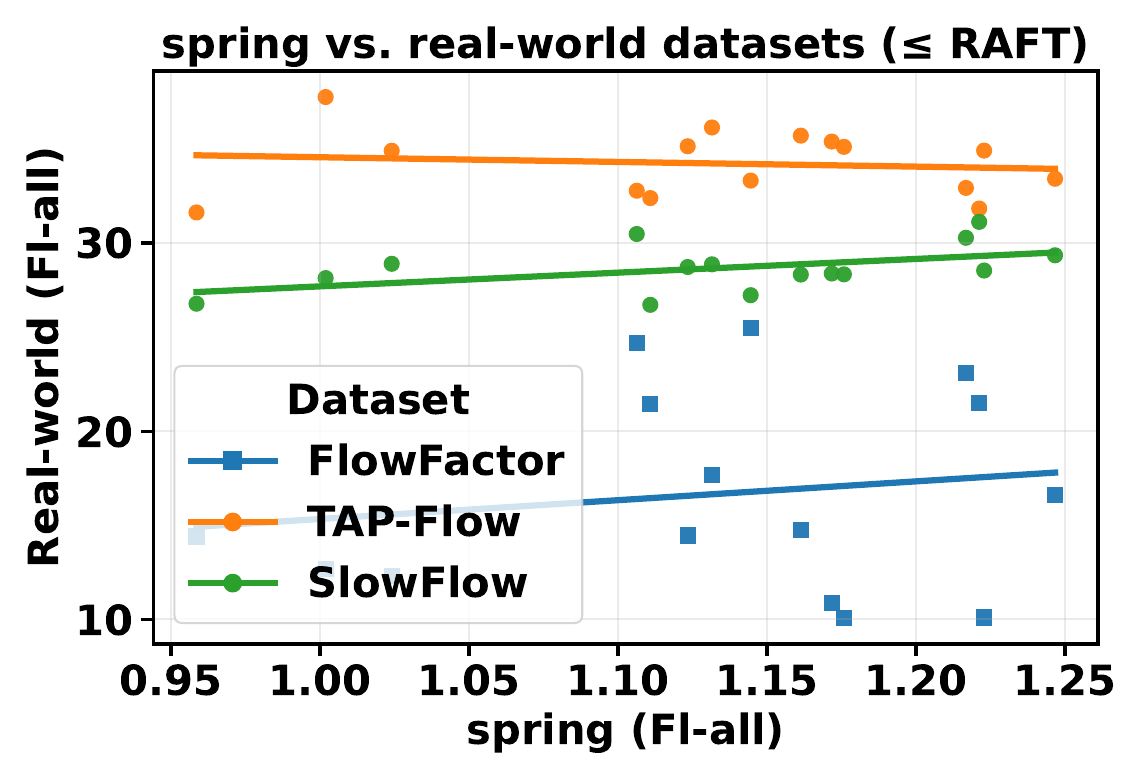}
  \caption{\textbf{Correlation between real-world datasets and KITTI,  Sintel-Final and Spring for models outperforming RAFT.}
  We plot the Fl-error of every Things model on KITTI \cite{menze2015object} Sintel-Final \cite{butler2012naturalistic} and Spring \cite{mehl2023spring} training sets  against FlowFactor, Slow Flow \cite{janai2017slow} and TAP-Flow \cite{doersch2022tap}, restricting to models that surpass RAFT \cite{teed2020raft} on KITTI \cite{menze2015object}, Sintel \cite{butler2012naturalistic} and Spring \cite{mehl2023spring}. The trendline is the OLS fit. While newer architectures achieve lower errors on these benchmarks, real-world accuracy stagnates or even degrades. This suggests that below RAFT-level Fl-all errors, current benchmarks no longer reflect real-world progress, highlighting the need for stronger real-world generalisation benchmarks.}
  \label{fig:generalisability_OLS_raft}
\end{figure}

However, when we restrict our attention to more modern models that outperform the seminal RAFT~\cite{teed2020raft} method on the training sets of KITTI \cite{menze2015object}, Sintel \cite{butler2012naturalistic}, and Spring \cite{mehl2023spring} a different picture emerges: these stronger current-benchmark models tend to stagnate or even perform worse on real-world data (\cref{fig:generalisability_OLS_raft}). We choose RAFT as a reference point because after that more and more models have implemented attention-based features and increased model capacity, which has repeatedly been observed to increase the risk of overfitting on narrow synthetic distributions \cite{weng2023attention}. As the displacement distributions of FlyingChairs and Sintel are closely aligned \cite{mayer2018makes}, we hypothesise that higher accuracy on the Sintel training set further may come at the cost of accuracy on other motion distributions, such as those in our real-world benchmarks.

\begin{figure}[!t]
     \centering
     \begin{subfigure}[b]{0.24\textwidth}
         \centering
         \includegraphics[width=\textwidth]{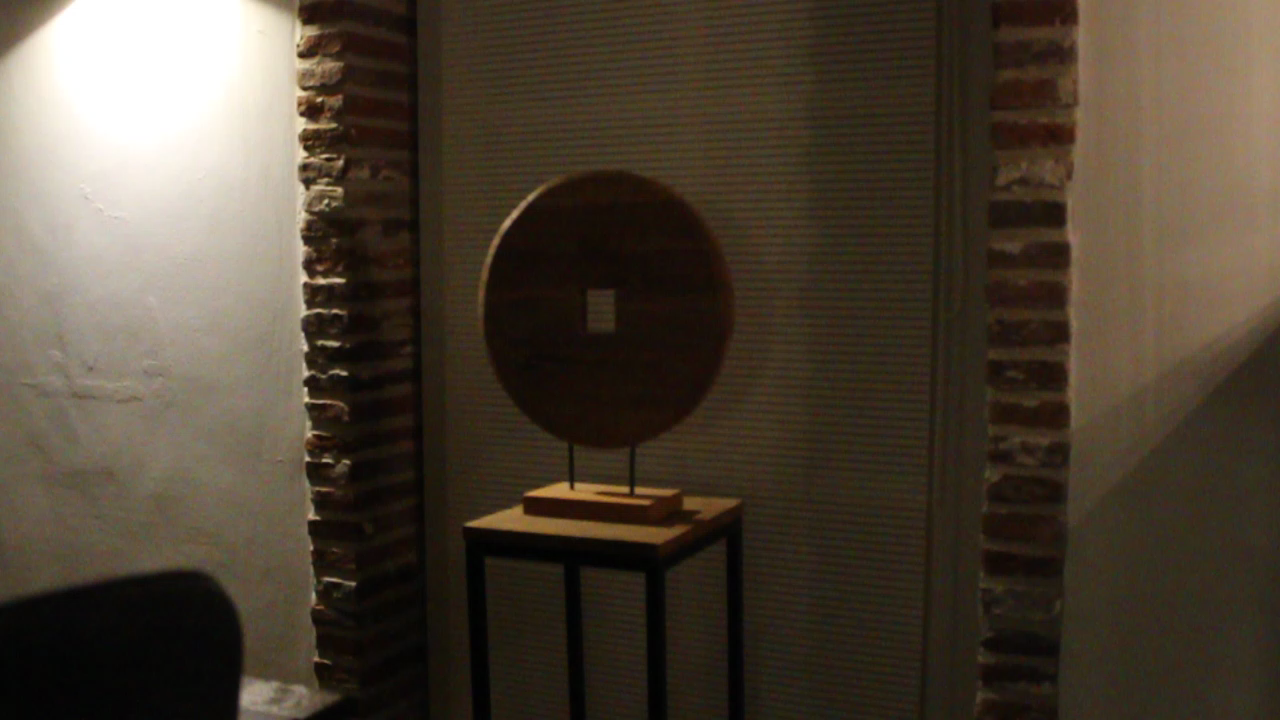}
         \caption*{Frame 1 (light cat.)}
     \end{subfigure}
     \hfill
     \begin{subfigure}[b]{0.24\textwidth}
         \centering
         \includegraphics[width=\textwidth]{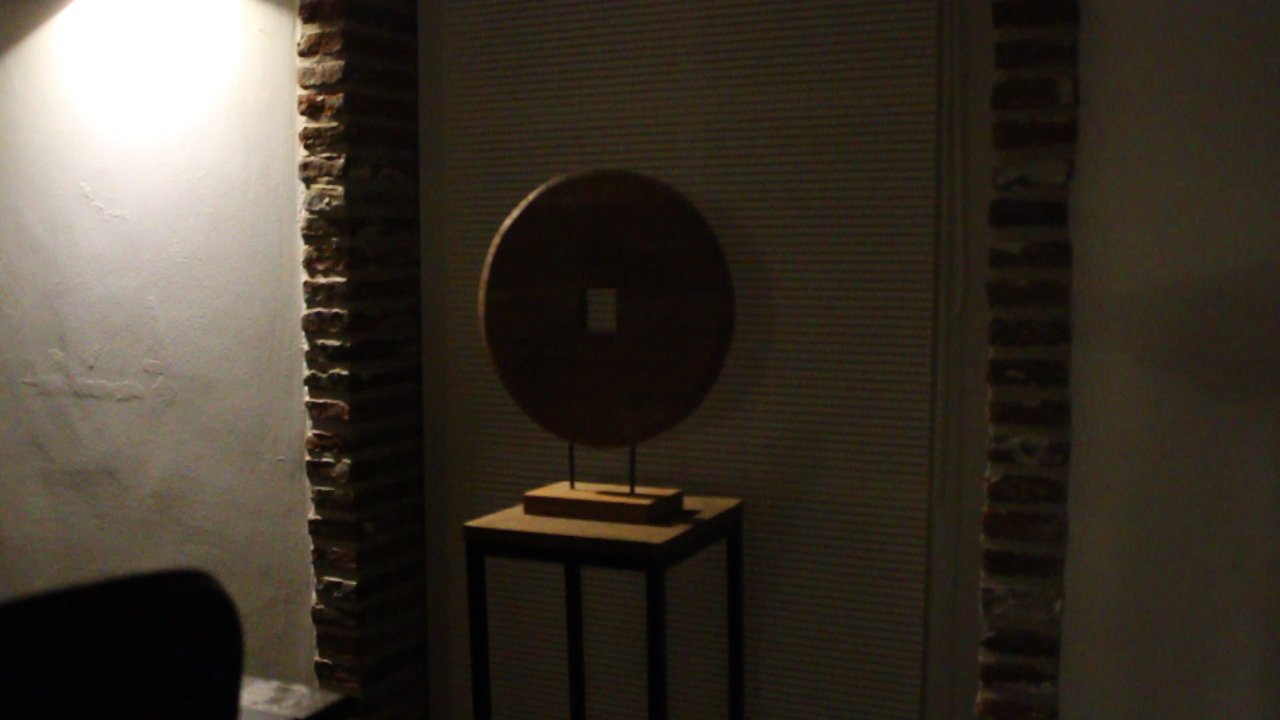}
         \caption*{Frame 2 (light cat.)}
     \end{subfigure}
     \hfill
     \begin{subfigure}[b]{0.24\textwidth}
         \centering
         \includegraphics[width=\textwidth]{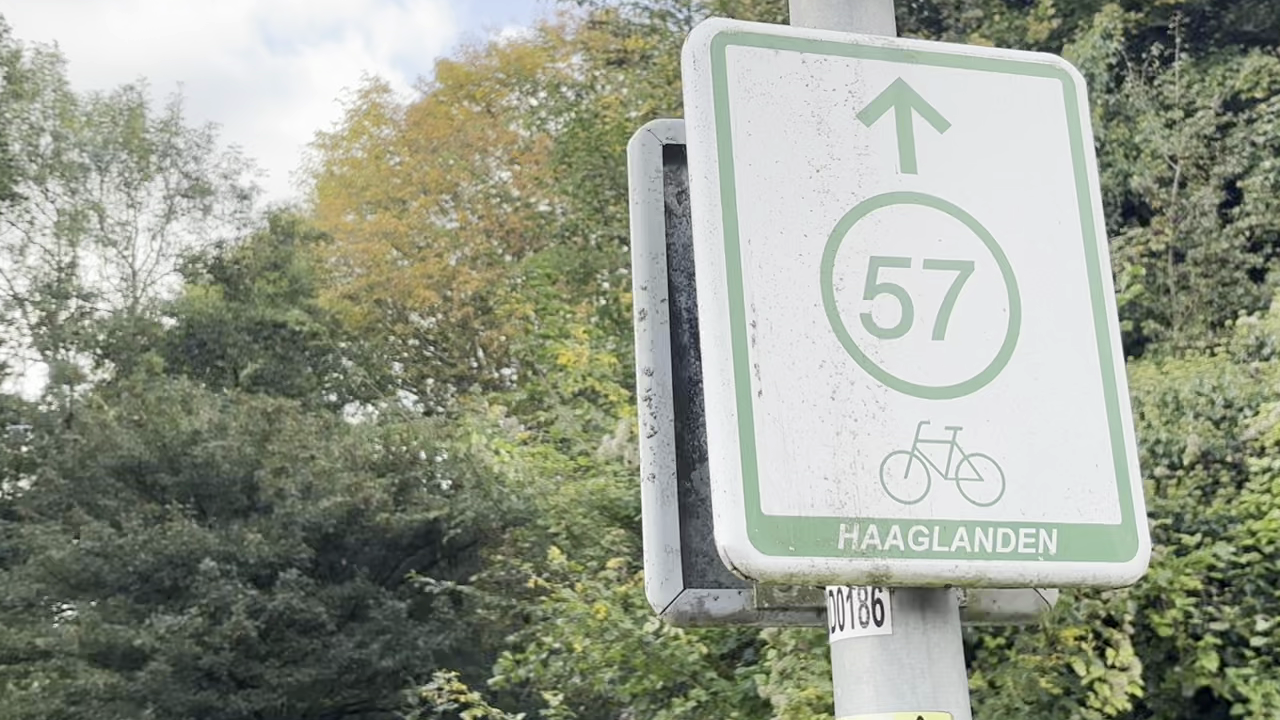}
         \caption*{Frame 1 (occlusion cat.)}
     \end{subfigure}
     \hfill
     \begin{subfigure}[b]{0.24\textwidth}
         \centering
         \includegraphics[width=\textwidth]{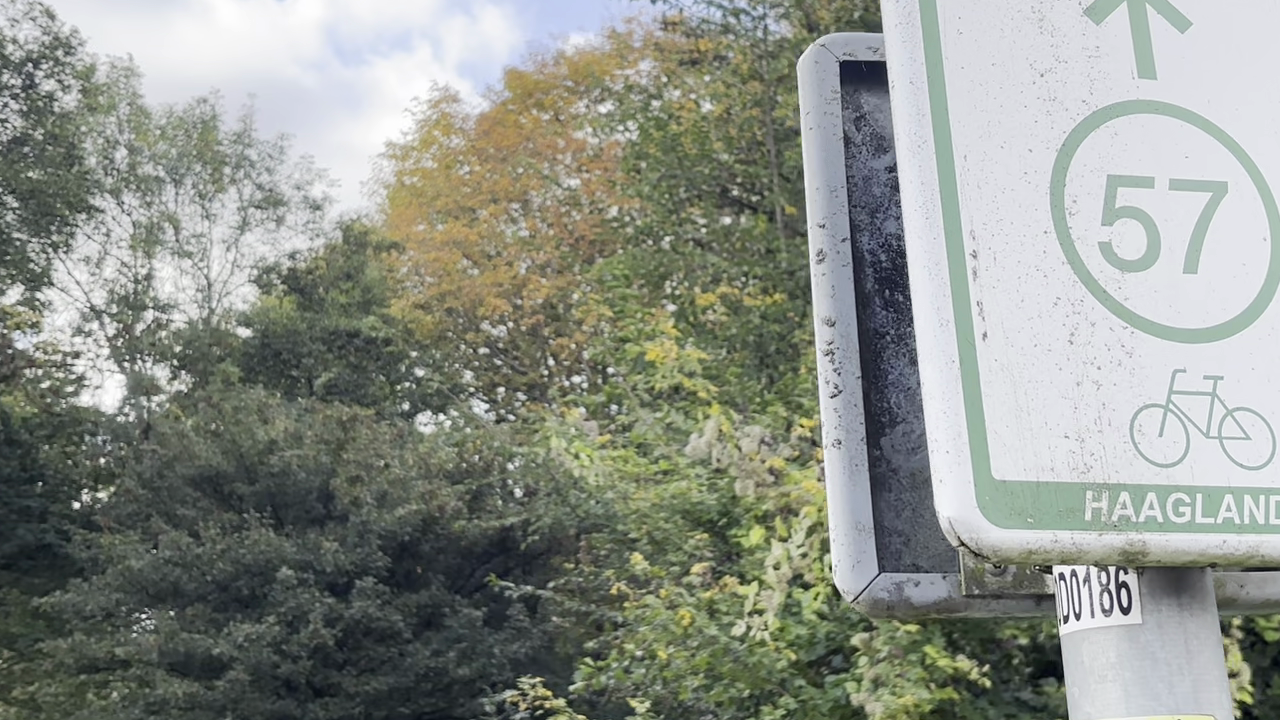}
         \caption*{Frame 2 (occlusion cat.)}
     \end{subfigure}

     \vspace{1pt} 

     \begin{subfigure}[b]{0.24\textwidth}
         \centering
         \includegraphics[width=\textwidth]{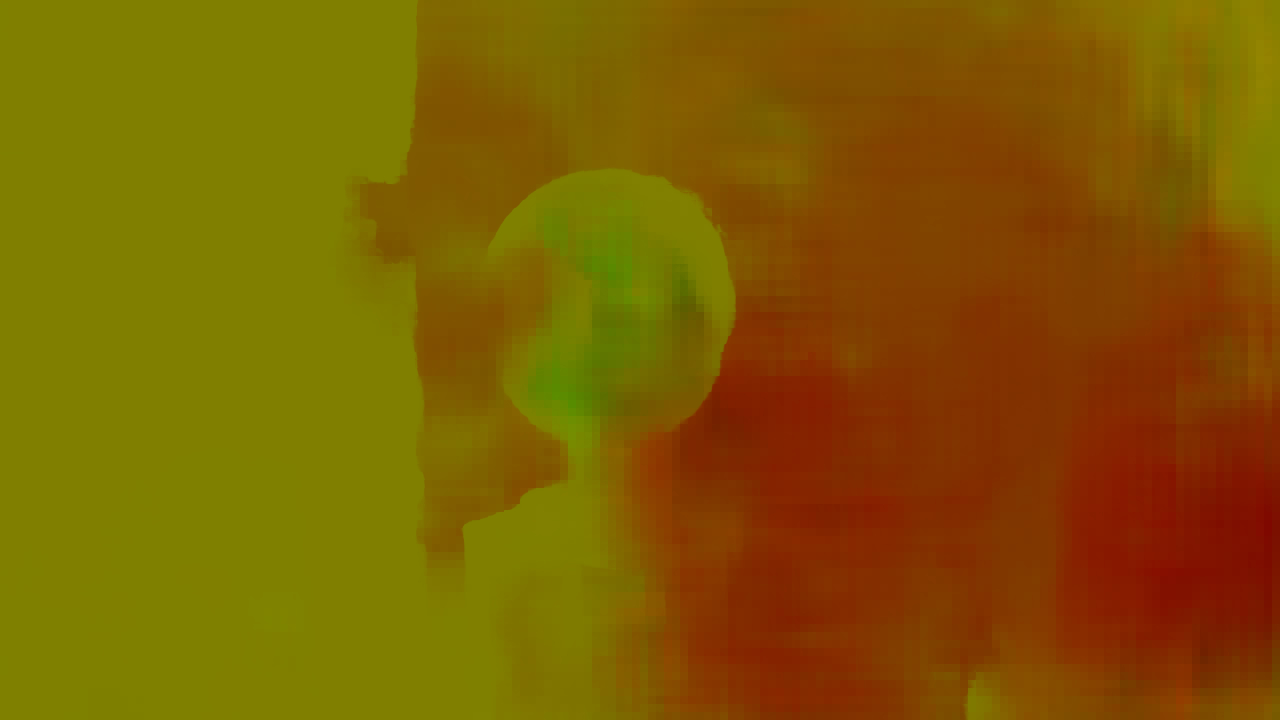}
         \caption*{SEA-RAFT \cite{wang2024sea} pred.}
     \end{subfigure}
     \hfill
     \begin{subfigure}[b]{0.24\textwidth}
         \centering
         \includegraphics[width=\textwidth]{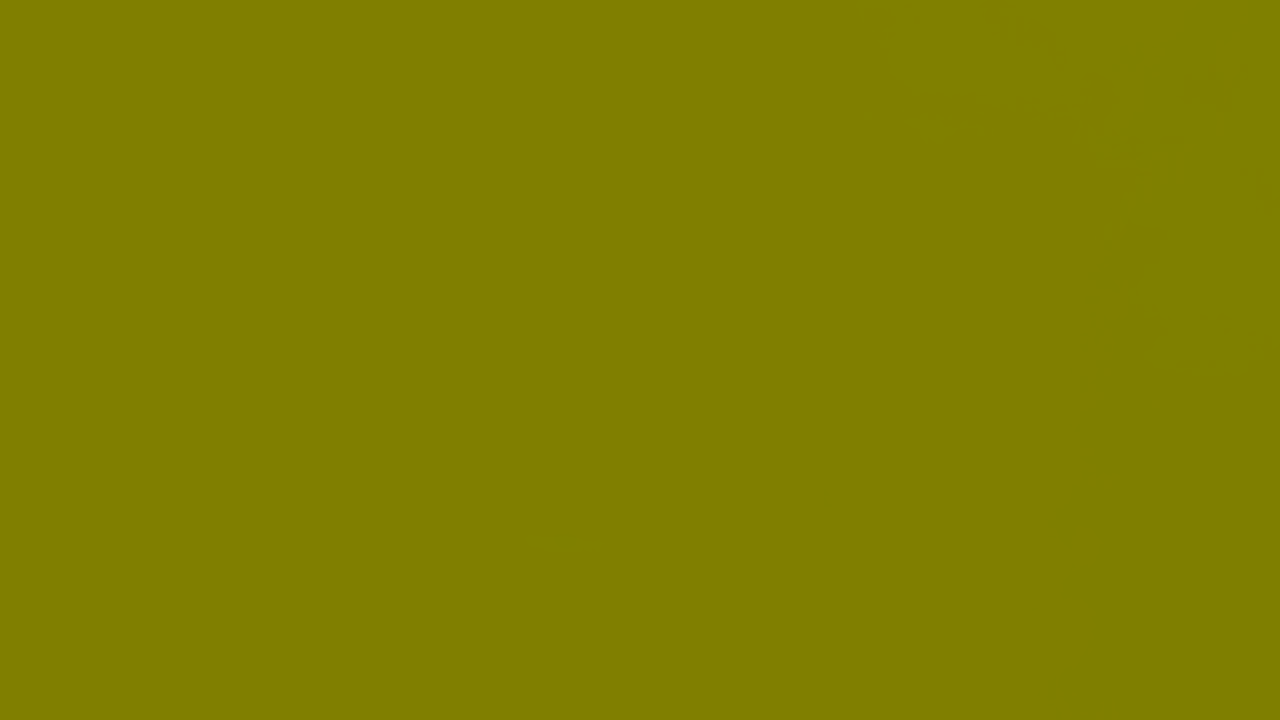}
         \caption*{GMA \cite{jiang2021learning} pred.}
     \end{subfigure}
     \hfill
     \begin{subfigure}[b]{0.24\textwidth}
         \centering
         \includegraphics[width=\textwidth]{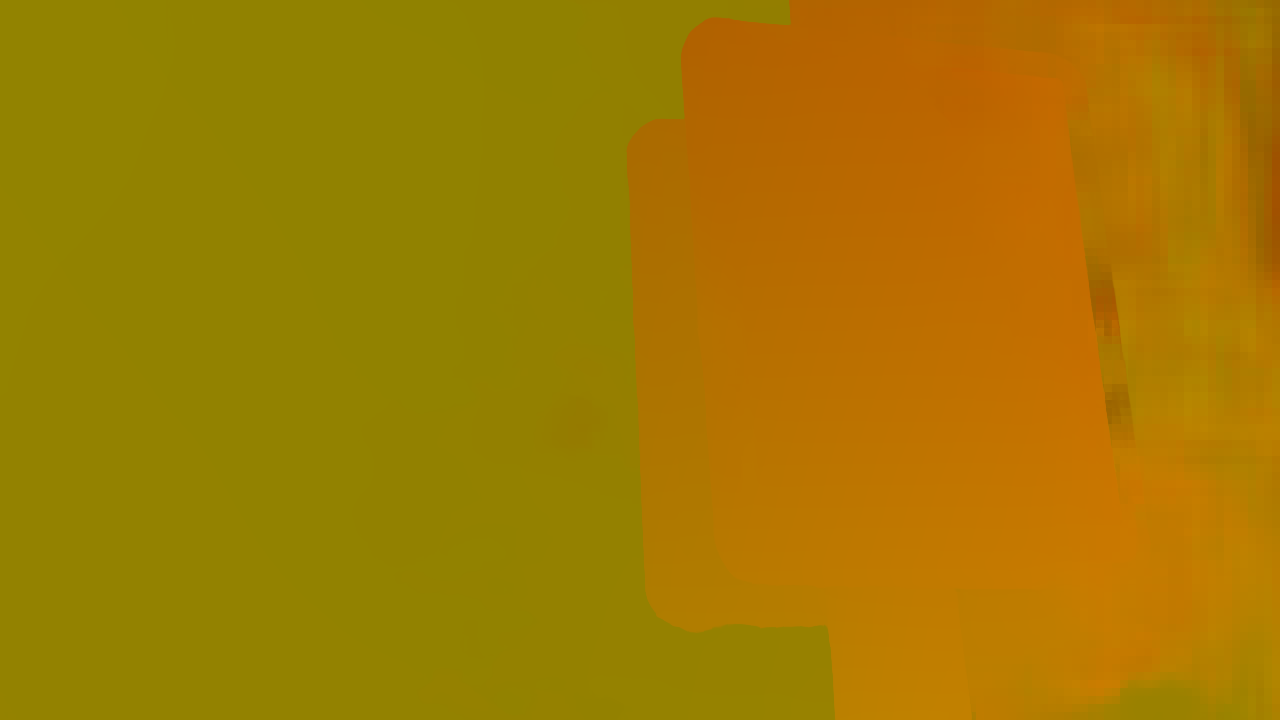}
         \caption*{SEA-RAFT \cite{wang2024sea} pred.}
     \end{subfigure}
     \hfill
     \begin{subfigure}[b]{0.24\textwidth}
         \centering
         \includegraphics[width=\textwidth]{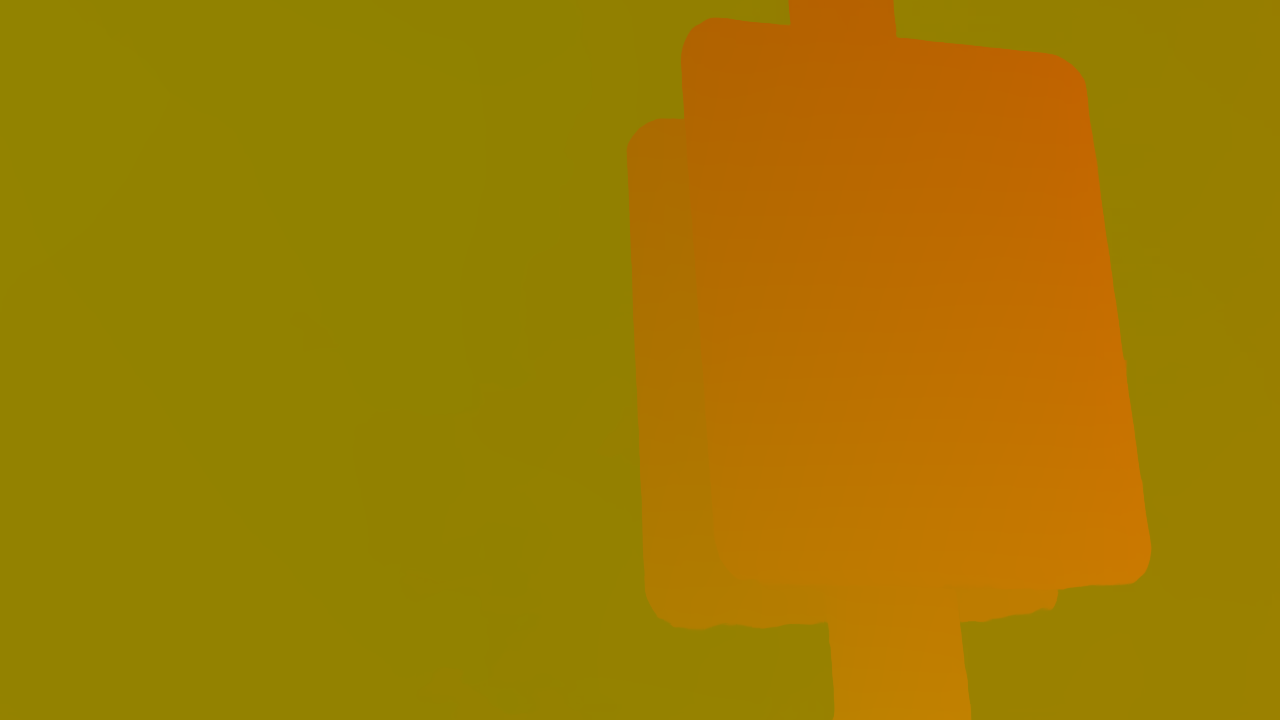}
         \caption*{GMA \cite{jiang2021learning} pred.}
     \end{subfigure}
     
     \caption{\textbf{Qualitative analysis.} The GMA model \cite{jiang2021learning} demonstrates superior stability under conditions of extreme photometric volatility. Specifically, it maintains a near-zero motion field when subjected to drastic lighting changes, whereas the SEA-RAFT small model \cite{wang2024sea} exhibits sensitivity to these variations, erroneously predicting motion. A similar performance gap is observed in the occlusion category. GMA \cite{jiang2021learning}: Robustly leverages global motion aggregation to mitigate the loss of local visual cues. SEA-RAFT small \cite{wang2024sea}: Undergoes significant performance degradation when spatial correspondence is interrupted. These discrepancies in real-world robustness are notably obscured by standard evaluation metrics. On the KITTI benchmark \cite{menze2015object}, the primary metric for assessing real-world generalisation, both GMA \cite{jiang2021learning} (C+T) and SEA-RAFT small \cite{wang2024sea} (C+T) yield comparable accuracy scores. This suggests that current benchmarks may lack the diversity required to expose the underlying vulnerabilities of certain architectures.}
     \label{fig:qualitative}
\end{figure}

Overall, we find that real-world accuracy has stagnated despite continued improvements on KITTI \cite{menze2015object}, Sintel \cite{butler2012naturalistic}, and Spring \cite{mehl2023spring}. This is consistent with Bauer \etal \cite{bauergeneralization}, who observed a similar trend for a mostly synthetic uncontrolled setting. These results call into question whether KITTI \cite{menze2015object}, Sintel \cite{butler2012naturalistic} and Spring \cite{mehl2023spring} are still sufficient markers of generalisability: are the latest reductions in error on these benchmarks indicative of real-world progress, or do they mainly reflect increased specialisation to these domains?
In~\cref{fig:qualitative} we show a qualitative comparison between models, where we show that SEA-RAFT \cite{wang2024sea}, a model that scores strongly on benchmarks such as Spring, fails to generalise to the real world. 

\subsection{Exp 2: isolated categories vs  real-world datasets}
\begin{figure}
\centering
    \includegraphics[width=0.4\columnwidth]{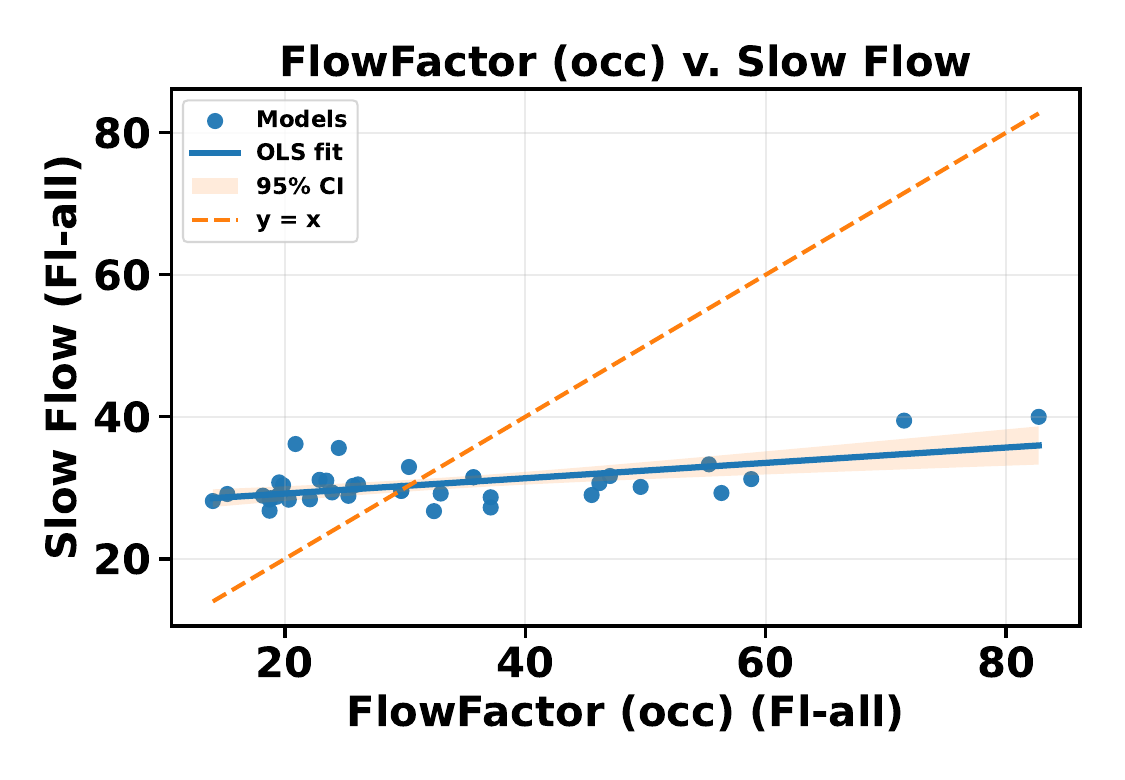}%
    \includegraphics[width=0.4\columnwidth]{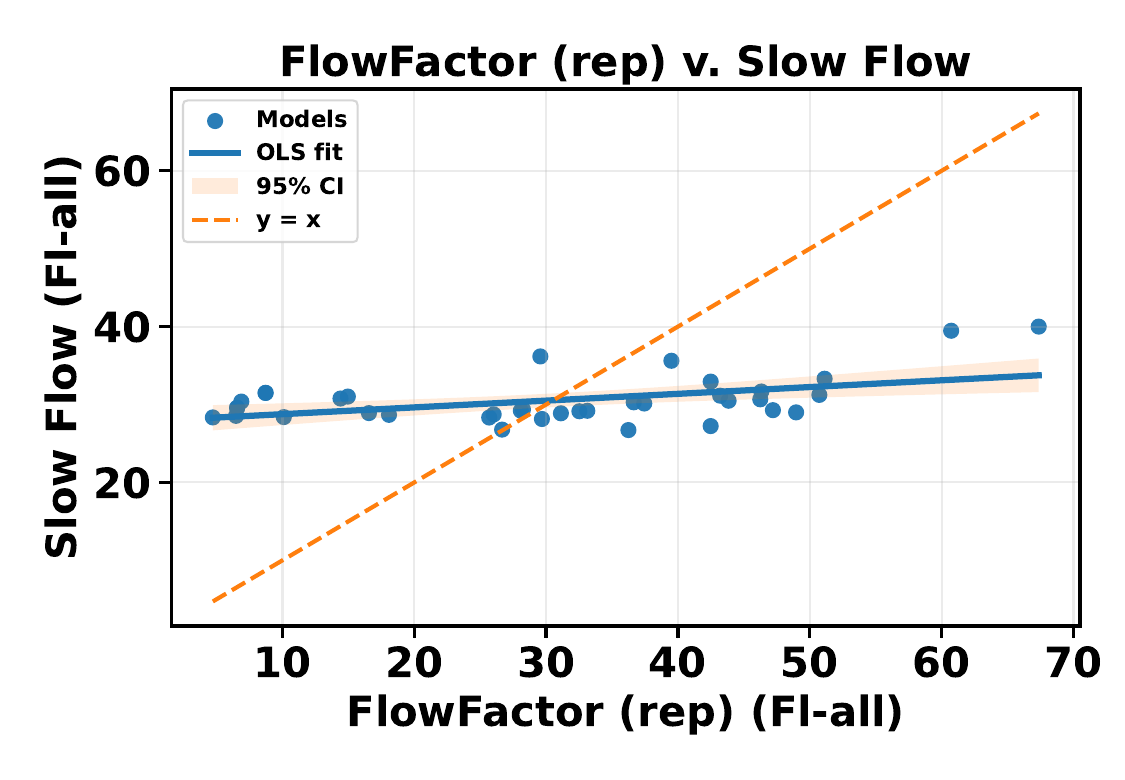}
    \includegraphics[width=0.4\columnwidth]{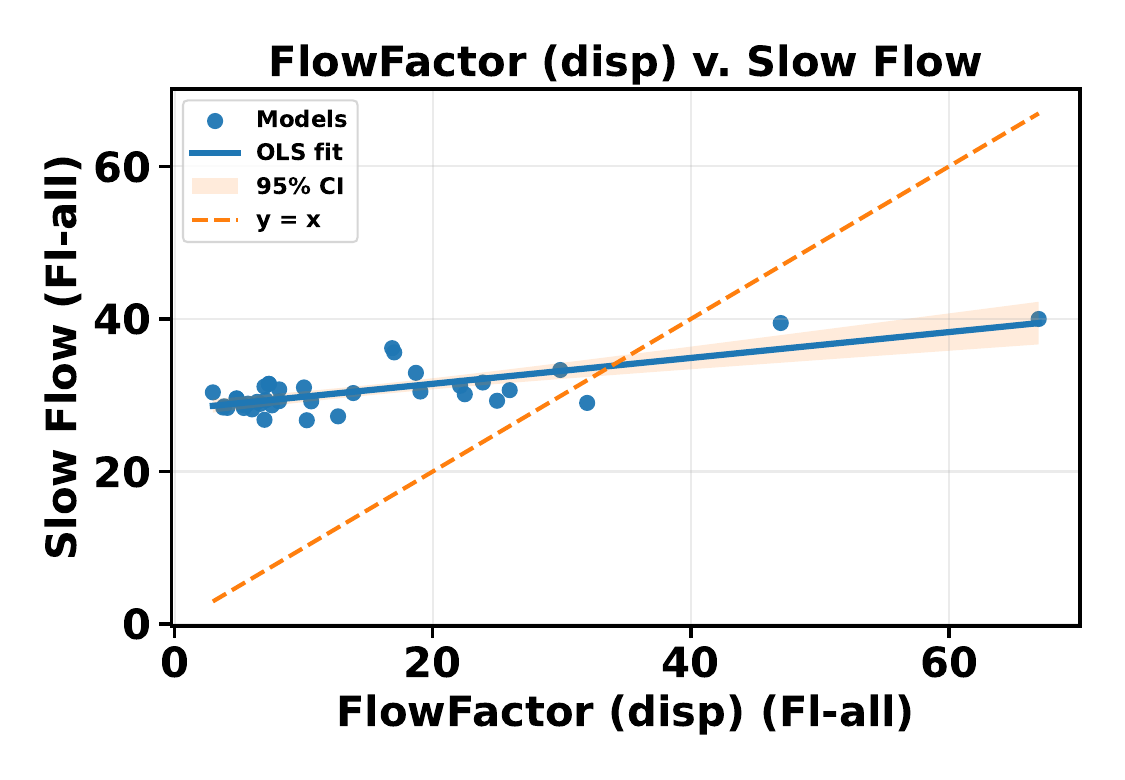}%
    \includegraphics[width=0.4\columnwidth]{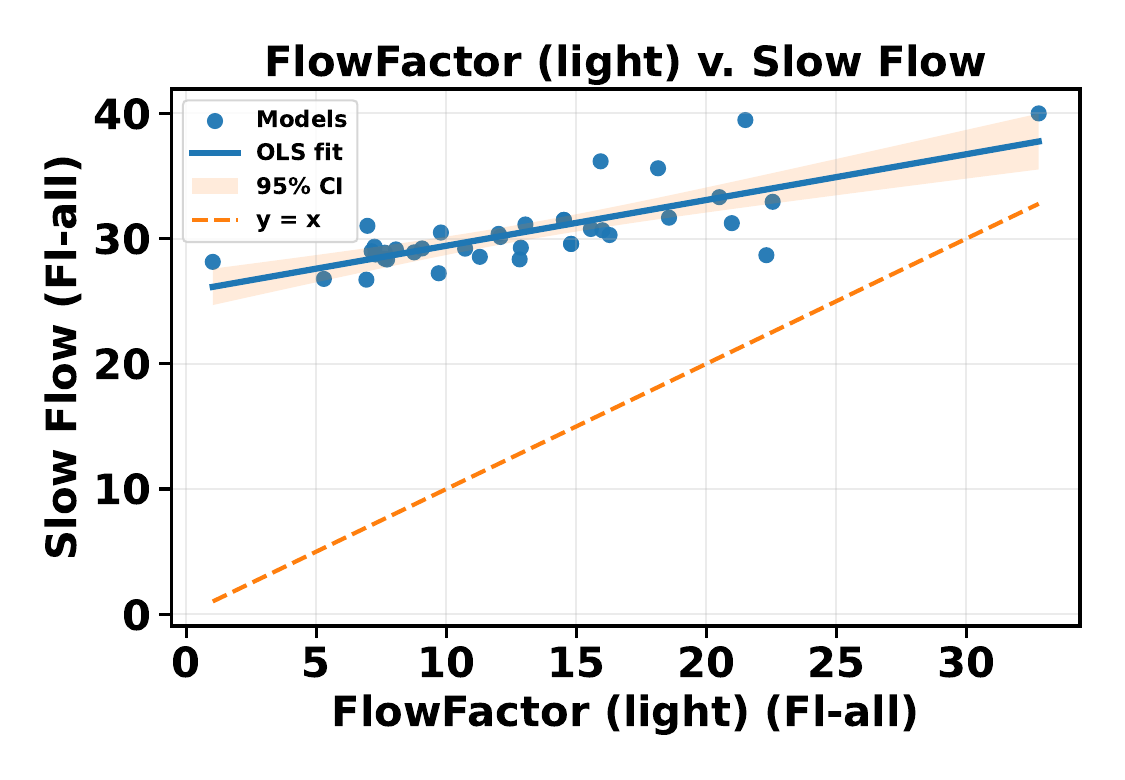}
  \caption{\textbf{Correlation between Slow Flow and isolated categories in FlowFactor.}
  Each panel shows the correlation between accuracy on a single FlowFactor category and accuracy on Slow Flow \cite{janai2017slow}. All categories exhibit positive correlation, but lighting ($\rho = 0.75$) and large displacements ($\rho = 0.74$) correlate most strongly, suggesting that improving robustness to these factors is particularly important for real-world accuracy.}
  \label{fig:corr_slow_factor}
\end{figure}

We next investigate which factors in the FlowFactor dataset most strongly correlate with real-world accuracy on Slow Flow \cite{janai2017slow}. In~\cref{fig:corr_slow_factor} we show the Spearman rank correlation coefficient between accuracy on each isolated FlowFactor category and accuracy on Slow Flow \cite{janai2017slow}. The varying-lighting category and the large-displacement category exhibit the strongest positive correlations with Slow Flow accuracy, indicating that these two factors are particularly predictive of real-world behaviour. Notice that some models deviate from the trend as they tend to overfit on certain aspects. We show that in general, all our categories positively correlate with real-world accuracy. We also show that the confounders correlate similarly with the TAP-Flow dataset; this can be found in the supplementary material. 

\subsection{Exp. 3: Inter-correlation of isolated categories}
\begin{figure}
  \centering
  \includegraphics[width=0.4\linewidth]{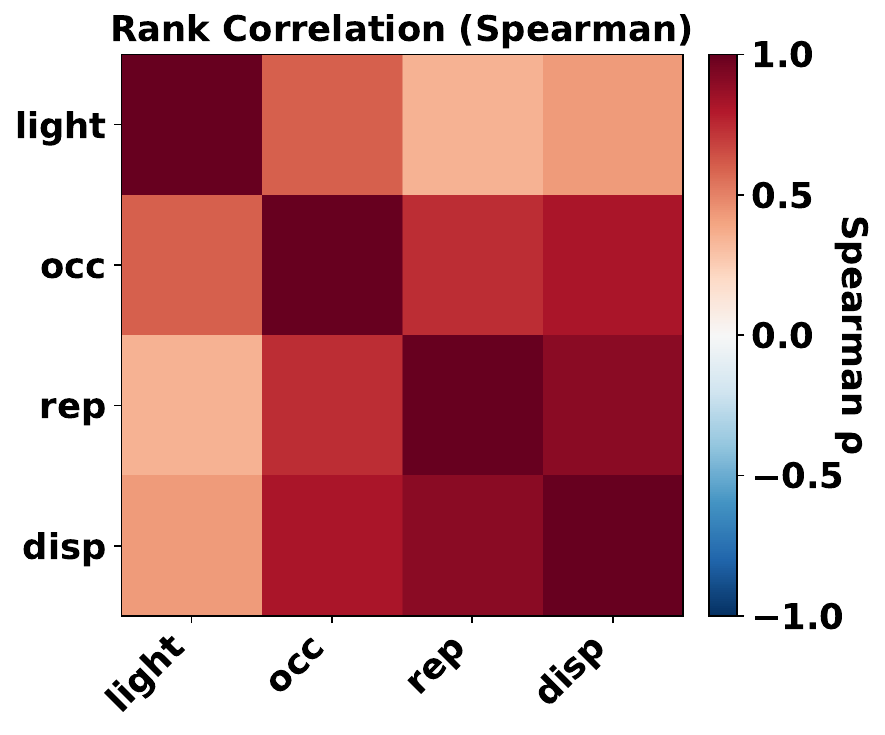}%
  \includegraphics[width=0.4\linewidth]{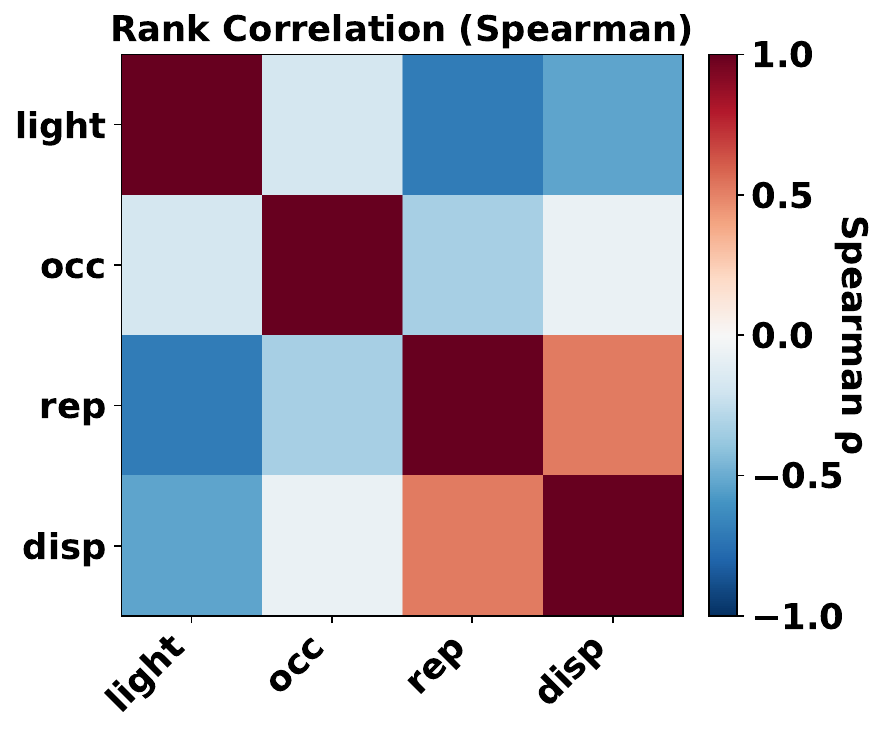}
  \caption{\textbf{Correlation between rankings on individual categories of the FlowFactor dataset.}
  Spearman rank correlation between model rankings on each pair of FlowFactor categories. Left: all models. Right: restricted to models that appear in the top-15 for at least one category, highlighting emerging trade-offs between displacement-heavy and stationary, distortion-based factors. Showing that there is no clear model, that works in every scenario.}
  \label{fig:rank_corrs}
\end{figure}

To study how different factors interact across models, we compute the Spearman rank correlation between model rankings for each pair of FlowFactor categories. This allows us to understand whether models that perform well on one factor tend to perform well or poorly on another. The results are shown in~\cref{fig:rank_corrs}.

The left heatmap uses all models and reveals that, globally, categories are positively correlated: models that are strong on one factor tend to be strong on others as well. The right heatmap is restricted to models that appear in the top-10 for at least one category. In this high-performing regime, we observe that the displacement- and repetitive-texture categories correlate negatively with the lighting and occlusion categories. This suggests a trade-off: models that are highly optimised for large displacements, as in the repetitive textures and large displacements category, tend to perform worse on the small-displacement, near-stationary scenes used for lighting and occlusion. In other words, robustness to photometric distortions in stationary scenes can conflict with challenging motion patterns, hinting at an inherent tension between robustness and accuracy.

\subsection{Exp. 4: To what extent does extra training data reduce the generalisation gap to the real world?}
\begin{figure}
  \centering
  \includegraphics[width=1.0\linewidth]{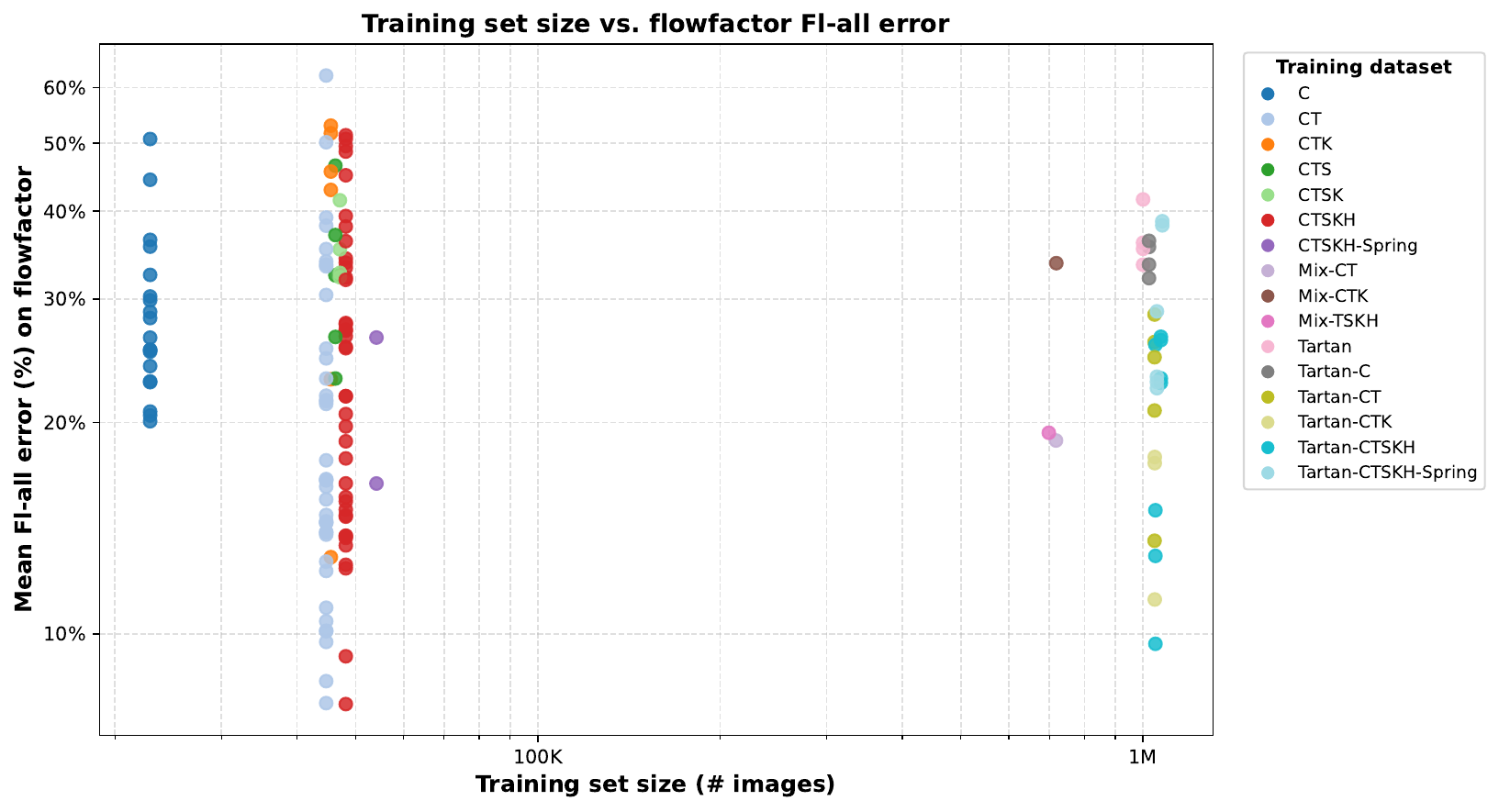}%
  \caption{\textbf{Size training dataset vs. accuracy on FlowFactor.}
  Here we plot for all models all possible checkpoints. For every checkpoint we plot the accompanying training dataset size and accuracy on our real-world dataset FlowFactor. Increasing the pre-training dataset size with more synthetic data, does not necessarily improve accuracy on the FlowFactor dataset, compared to C+T training.}
  \label{fig:train_size_flowfactor}
\end{figure}

Throughout this paper we have used the same C+T things checkpoints to keep a fair comparison between models. However, in the real-world one could use as much data as one would like. Recently, models have started to incorporate the synthetic dataset TartanAir \cite{wang2020tartanair} into their pre-training regime \cite{wang2024sea, zhang2026megaflow, wang2025waft}, which contains more than 1 million frame pairs. Others have changed the entire protocol by now mixing different labeled and unlabeled datasets into their training regime \cite{liang2025flow}. Does this increase in pre-training dataset size help with generalisation to the real-world? In~\cref{fig:train_size_flowfactor} we show that extra synthetic data on top of C+T does not necessarily improve performance. The models sporting the highest accuracy on FlowFactor use either C+T or C+T in combination with a small amount of more realistic data (C+T+S+K+H). We show a similar pattern for the other 2 datasets in the supplementary material.  

\section{Conclusion}
We created a factorised real-world dataset containing 1,000 frame pairs. We showed that the generalisability of recent models to the real world has stagnated and that current benchmarks do not measure real-world accuracy. We further demonstrated that accuracy across all factors in our factorised dataset correlates positively with accuracy on real-world benchmarks. In addition, we found that high accuracy on large displacements and repetitive textures correlates negatively with accuracy under varying lighting conditions and occlusions. Furthermore, we show that training on extra synthetic data does not necessarily reduce the generalisability gap to the real world. \\

\noindent
\textbf{Acknowledgements}
This project is supported by NWO (project VI.C.242.088). We would like to thank the following people for their contributions: Iris Petre, Jesse Klijnsma, Zhuoyue Ge, Marijn Timmerije, and Sachhyam Dahal. Furthermore, we would like to thank the 5 people that participated in the annotation survey.  \\

\noindent
\textbf{Environmental statement}
During this study, an NVIDIA A40 was used for approximately 553.5 hours, resulting in an estimated 44.86 kg $CO_2$-equivalent. These emissions were offset via certified (re)forestation credits \footnote{\url{https://treesforall.nl/app/uploads/certificates/Certificaat_Trees_for_All_1615061d.pdf}}.

%
%
\bibliographystyle{splncs04}
\bibliography{main}

@article{schmalfuss2025robustspring,
  title={Robustspring: Benchmarking robustness to image corruptions for optical flow, scene flow and stereo},
  author={Schmalfuss, Jenny and Oei, Victor and Mehl, Lukas and Bartsch, Madlen and Agnihotri, Shashank and Keuper, Margret and Bruhn, Andr{\'e}s},
  journal={arXiv preprint arXiv:2505.09368},
  year={2025}
}

@article{liang2025flow,
  title={Flow-Anything: Learning Real-World Optical Flow Estimation from Large-Scale Single-view Images},
  author={Liang, Yingping and Fu, Ying and Hu, Yutao and Shao, Wenqi and Liu, Jiaming and Zhang, Debing},
  journal={arXiv preprint arXiv:2506.07740},
  year={2025}
}

@inproceedings{menze2015object,
  title={Object scene flow for autonomous vehicles},
  author={Menze, Moritz and Geiger, Andreas},
  booktitle={Proceedings of the IEEE conference on computer vision and pattern recognition},
  pages={3061--3070},
  year={2015}
}

@inproceedings{butler2012naturalistic,
  title={A naturalistic open source movie for optical flow evaluation},
  author={Butler, Daniel J and Wulff, Jonas and Stanley, Garrett B and Black, Michael J},
  booktitle={European conference on computer vision},
  pages={611--625},
  year={2012},
  organization={Springer}
}

@inproceedings{mehl2023spring,
  title={Spring: A high-resolution high-detail dataset and benchmark for scene flow, optical flow and stereo},
  author={Mehl, Lukas and Schmalfuss, Jenny and Jahedi, Azin and Nalivayko, Yaroslava and Bruhn, Andr{\'e}s},
  booktitle={Proceedings of the IEEE/CVF Conference on Computer Vision and Pattern Recognition},
  pages={4981--4991},
  year={2023}
}

@inproceedings{zheng2020optical,
  title={Optical flow in the dark},
  author={Zheng, Yinqiang and Zhang, Mingfang and Lu, Feng},
  booktitle={Proceedings of the IEEE/CVF conference on computer vision and pattern recognition},
  pages={6749--6757},
  year={2020}
}

@inproceedings{greff2022kubric,
  title={Kubric: A scalable dataset generator},
  author={Greff, Klaus and Belletti, Francois and Beyer, Lucas and Doersch, Carl and Du, Yilun and Duckworth, Daniel and Fleet, David J and Gnanapragasam, Dan and Golemo, Florian and Herrmann, Charles and others},
  booktitle={Proceedings of the IEEE/CVF conference on computer vision and pattern recognition},
  pages={3749--3761},
  year={2022}
}

@article{zhai2021optical,
  title={Optical flow and scene flow estimation: A survey},
  author={Zhai, Mingliang and Xiang, Xuezhi and Lv, Ning and Kong, Xiangdong},
  journal={Pattern Recognition},
  volume={114},
  pages={107861},
  year={2021},
  publisher={Elsevier}
}

@inproceedings{sun2021autoflow,
  title={Autoflow: Learning a better training set for optical flow},
  author={Sun, Deqing and Vlasic, Daniel and Herrmann, Charles and Jampani, Varun and Krainin, Michael and Chang, Huiwen and Zabih, Ramin and Freeman, William T and Liu, Ce},
  booktitle={Proceedings of the IEEE/CVF Conference on Computer Vision and Pattern Recognition},
  pages={10093--10102},
  year={2021}
}

@article{mayer2018makes,
  title={What makes good synthetic training data for learning disparity and optical flow estimation?},
  author={Mayer, Nikolaus and Ilg, Eddy and Fischer, Philipp and Hazirbas, Caner and Cremers, Daniel and Dosovitskiy, Alexey and Brox, Thomas},
  journal={International Journal of Computer Vision},
  volume={126},
  number={9},
  pages={942--960},
  year={2018},
  publisher={Springer}
}

@inproceedings{dosovitskiy2015flownet,
  title={Flownet: Learning optical flow with convolutional networks},
  author={Dosovitskiy, Alexey and Fischer, Philipp and Ilg, Eddy and Hausser, Philip and Hazirbas, Caner and Golkov, Vladimir and Van Der Smagt, Patrick and Cremers, Daniel and Brox, Thomas},
  booktitle={Proceedings of the IEEE international conference on computer vision},
  pages={2758--2766},
  year={2015}
}

@inproceedings{sun2018pwc,
  title={Pwc-net: Cnns for optical flow using pyramid, warping, and cost volume},
  author={Sun, Deqing and Yang, Xiaodong and Liu, Ming-Yu and Kautz, Jan},
  booktitle={Proceedings of the IEEE conference on computer vision and pattern recognition},
  pages={8934--8943},
  year={2018}
}

@inproceedings{teed2020raft,
  title={Raft: Recurrent all-pairs field transforms for optical flow},
  author={Teed, Zachary and Deng, Jia},
  booktitle={Computer Vision--ECCV 2020: 16th European Conference, Glasgow, UK, August 23--28, 2020, Proceedings, Part II 16},
  pages={402--419},
  year={2020},
  organization={Springer}
}

@inproceedings{jiang2021learning,
  title={Learning to estimate hidden motions with global motion aggregation},
  author={Jiang, Shihao and Campbell, Dylan and Lu, Yao and Li, Hongdong and Hartley, Richard},
  booktitle={Proceedings of the IEEE/CVF international conference on computer vision},
  pages={9772--9781},
  year={2021}
}

@inproceedings{huang2022flowformer,
  title={Flowformer: A transformer architecture for optical flow},
  author={Huang, Zhaoyang and Shi, Xiaoyu and Zhang, Chao and Wang, Qiang and Cheung, Ka Chun and Qin, Hongwei and Dai, Jifeng and Li, Hongsheng},
  booktitle={European conference on computer vision},
  pages={668--685},
  year={2022},
  organization={Springer}
}

@inproceedings{shi2023flowformer++,
  title={Flowformer++: Masked cost volume autoencoding for pretraining optical flow estimation},
  author={Shi, Xiaoyu and Huang, Zhaoyang and Li, Dasong and Zhang, Manyuan and Cheung, Ka Chun and See, Simon and Qin, Hongwei and Dai, Jifeng and Li, Hongsheng},
  booktitle={Proceedings of the IEEE/CVF conference on computer vision and pattern recognition},
  pages={1599--1610},
  year={2023}
}

@inproceedings{xu2022gmflow,
  title={Gmflow: Learning optical flow via global matching},
  author={Xu, Haofei and Zhang, Jing and Cai, Jianfei and Rezatofighi, Hamid and Tao, Dacheng},
  booktitle={Proceedings of the IEEE/CVF conference on computer vision and pattern recognition},
  pages={8121--8130},
  year={2022}
}

@inproceedings{zhao2022global,
  title={Global matching with overlapping attention for optical flow estimation},
  author={Zhao, Shiyu and Zhao, Long and Zhang, Zhixing and Zhou, Enyu and Metaxas, Dimitris},
  booktitle={Proceedings of the IEEE/CVF Conference on Computer Vision and Pattern Recognition},
  pages={17592--17601},
  year={2022}
}

@inproceedings{ilg2017flownet,
  title={Flownet 2.0: Evolution of optical flow estimation with deep networks},
  author={Ilg, Eddy and Mayer, Nikolaus and Saikia, Tonmoy and Keuper, Margret and Dosovitskiy, Alexey and Brox, Thomas},
  booktitle={Proceedings of the IEEE conference on computer vision and pattern recognition},
  pages={2462--2470},
  year={2017}
}

@inproceedings{eslami2024rethinking,
  title={Rethinking raft for efficient optical flow},
  author={Eslami, Navid and Arefi, Farnoosh and Mansourian, Amir M and Kasaei, Shohreh},
  booktitle={2024 13th Iranian/3rd International Machine Vision and Image Processing Conference (MVIP)},
  pages={1--7},
  year={2024},
  organization={IEEE}
}

@inproceedings{mayer2016large,
  title={A large dataset to train convolutional networks for disparity, optical flow, and scene flow estimation},
  author={Mayer, Nikolaus and Ilg, Eddy and Hausser, Philip and Fischer, Philipp and Cremers, Daniel and Dosovitskiy, Alexey and Brox, Thomas},
  booktitle={Proceedings of the IEEE conference on computer vision and pattern recognition},
  pages={4040--4048},
  year={2016}
}

@article{agnihotri2025flowbench,
  title={FlowBench: Benchmarking Optical Flow Estimation Methods for Reliability and Generalization},
  author={Agnihotri, Shashank and Caspary, Julian Yuya and Schwarz, Luca and Gao, Xinyan and Schmalfuss, Jenny and Bruhn, Andres and Keuper, Margret},
  journal={Transactions on Machine Learning Research},
  year={2025}
}

@article{bauergeneralization,
  title={On the Generalization of Optical Flow: Quantifying Robustness to Dataset Shifts},
  author={Bauer, Katrin and Bruhn, Andr{\'e}s and Schmalfuss, Jenny}
}

@inproceedings{liu2008human,
  title={Human-assisted motion annotation},
  author={Liu, Ce and Freeman, William T and Adelson, Edward H and Weiss, Yair},
  booktitle={2008 IEEE Conference on Computer Vision and Pattern Recognition},
  pages={1--8},
  year={2008},
  organization={IEEE}
}

@article{zhou2024adverse,
  title={Adverse weather optical flow: Cumulative homogeneous-heterogeneous adaptation},
  author={Zhou, Hanyu and Chang, Yi and Shi, Zhiwei and Yan, Wending and Chen, Gang and Tian, Yonghong and Yan, Luxin},
  journal={IEEE Transactions on Pattern Analysis and Machine Intelligence},
  year={2024},
  publisher={IEEE}
}

@article{doersch2022tap,
  title={Tap-vid: A benchmark for tracking any point in a video},
  author={Doersch, Carl and Gupta, Ankush and Markeeva, Larisa and Recasens, Adria and Smaira, Lucas and Aytar, Yusuf and Carreira, Joao and Zisserman, Andrew and Yang, Yi},
  journal={Advances in Neural Information Processing Systems},
  volume={35},
  pages={13610--13626},
  year={2022}
}

@inproceedings{li2021gyroflow,
  title={GyroFlow: Gyroscope-guided unsupervised optical flow learning},
  author={Li, Haipeng and Luo, Kunming and Liu, Shuaicheng},
  booktitle={Proceedings of the IEEE/CVF international conference on computer vision},
  pages={12869--12878},
  year={2021}
}

@inproceedings{weinzaepfel2015learning,
  title={Learning to detect motion boundaries},
  author={Weinzaepfel, Philippe and Revaud, Jerome and Harchaoui, Zaid and Schmid, Cordelia},
  booktitle={Proceedings of the IEEE conference on computer vision and pattern recognition},
  pages={2578--2586},
  year={2015}
}

@article{sturzel2004perceptual,
  title={Perceptual limits of common fate},
  author={St{\"u}rzel, Frank and Spillmann, Lothar},
  journal={Vision research},
  volume={44},
  number={13},
  pages={1565--1573},
  year={2004},
  publisher={Elsevier}
}

@article{baker2011database,
  title={A database and evaluation methodology for optical flow},
  author={Baker, Simon and Scharstein, Daniel and Lewis, James P and Roth, Stefan and Black, Michael J and Szeliski, Richard},
  journal={International journal of computer vision},
  volume={92},
  number={1},
  pages={1--31},
  year={2011},
  publisher={Springer}
}

@inproceedings{janai2017slow,
  title={Slow flow: Exploiting high-speed cameras for accurate and diverse optical flow reference data},
  author={Janai, Joel and Guney, Fatma and Wulff, Jonas and Black, Michael J and Geiger, Andreas},
  booktitle={Proceedings of the IEEE Conference on Computer Vision and Pattern Recognition},
  pages={3597--3607},
  year={2017}
}

@misc{morimitsu2021ptlflow,
  author = {Henrique Morimitsu},
  title = {PyTorch Lightning Optical Flow},
  year = {2021},
  publisher = {GitHub},
  journal = {GitHub repository},
  howpublished = {\url{https://github.com/hmorimitsu/ptlflow}}
}

@inproceedings{yin2019hierarchical,
  title={Hierarchical discrete distribution decomposition for match density estimation},
  author={Yin, Zhichao and Darrell, Trevor and Yu, Fisher},
  booktitle={Proceedings of the IEEE/CVF conference on computer vision and pattern recognition},
  pages={6044--6053},
  year={2019}
}

@inproceedings{zhao2020maskflownet,
  title={Maskflownet: Asymmetric feature matching with learnable occlusion mask},
  author={Zhao, Shengyu and Sheng, Yilun and Dong, Yue and Chang, Eric I and Xu, Yan and others},
  booktitle={Proceedings of the IEEE/CVF conference on computer vision and pattern recognition},
  pages={6278--6287},
  year={2020}
}

@article{weng2023attention,
  title={Attention mechanism trained with small datasets for biomedical image segmentation},
  author={Weng, Weihao and Zhu, Xin and Jing, Lei and Dong, Mianxiong},
  journal={Electronics},
  volume={12},
  number={3},
  pages={682},
  year={2023},
  publisher={MDPI}
}

@inproceedings{wang2020tartanair,
  title={Tartanair: A dataset to push the limits of visual slam},
  author={Wang, Wenshan and Zhu, Delong and Wang, Xiangwei and Hu, Yaoyu and Qiu, Yuheng and Wang, Chen and Hu, Yafei and Kapoor, Ashish and Scherer, Sebastian},
  booktitle={2020 IEEE/RSJ International Conference on Intelligent Robots and Systems (IROS)},
  pages={4909--4916},
  year={2020},
  organization={IEEE}
}

@article{alfarano2024estimating,
  title={Estimating optical flow: A comprehensive review of the state of the art},
  author={Alfarano, Andrea and Maiano, Luca and Papa, Lorenzo and Amerini, Irene},
  journal={Computer vision and image understanding},
  volume={249},
  pages={104160},
  year={2024},
  publisher={Elsevier}
}

@inproceedings{kondermann2016hci,
  title={The hci benchmark suite: Stereo and flow ground truth with uncertainties for urban autonomous driving},
  author={Kondermann, Daniel and Nair, Rahul and Honauer, Katrin and Krispin, Karsten and Andrulis, Jonas and Brock, Alexander and Gussefeld, Burkhard and Rahimimoghaddam, Mohsen and Hofmann, Sabine and Brenner, Claus and others},
  booktitle={Proceedings of the IEEE Conference on Computer Vision and Pattern Recognition Workshops},
  pages={19--28},
  year={2016}
}

@inproceedings{wang2024sea,
  title={Sea-raft: Simple, efficient, accurate raft for optical flow},
  author={Wang, Yihan and Lipson, Lahav and Deng, Jia},
  booktitle={European Conference on Computer Vision},
  pages={36--54},
  year={2024},
  organization={Springer}
}

@misc{robustvisionchallenge,
  title        = {Robust Vision Challenge},
  howpublished = {\url{http://www.robustvision.net/}},
  note         = {Accessed: 2025-11-13}
}

@inproceedings{geiger2012we,
  title={Are we ready for autonomous driving? the kitti vision benchmark suite},
  author={Geiger, Andreas and Lenz, Philip and Urtasun, Raquel},
  booktitle={2012 IEEE conference on computer vision and pattern recognition},
  pages={3354--3361},
  year={2012},
  organization={IEEE}
}

@article{zhang2026megaflow,
  title={MegaFlow: Zero-Shot Large Displacement Optical Flow},
  author={Zhang, Dingxi and Wang, Fangjinhua and Pollefeys, Marc and Xu, Haofei},
  journal={arXiv preprint arXiv:2603.25739},
  year={2026}
}

@article{wang2025waft,
  title={Waft: Warping-alone field transforms for optical flow},
  author={Wang, Yihan and Deng, Jia},
  journal={arXiv preprint arXiv:2506.21526},
  year={2025}
}

@thesis{Dahal2025,
  author      = {S. Dahal},
  title       = {Going Against The Flow: Evaluating Optical Flow Estimation Models on Real-World Non-Rigid Motion},
  type        = {Bachelor Thesis},
  school      = {Delft University of Technology},
  year        = {2025},
  url         = {https://resolver.tudelft.nl/uuid:ab2db1e7-9122-43c3-9377-f188fceb30b6}
}

@thesis{Timmerije2025,
  author      = {M. Timmerije},
  title       = {Bridging the Gap: A Real-World Dataset and Evaluation of Optical Flow Models in Large Displacement Scenarios},
  type        = {Bachelor Thesis},
  school      = {Delft University of Technology},
  year        = {2025},
  url         = {https://resolver.tudelft.nl/uuid:d7c4cc36-6582-47fc-94cd-cc0dc393725b}
}

@thesis{Ge2025,
  author      = {Z. Ge},
  title       = {Real-World Evaluation of Optical Flow with Varying Lighting Conditions},
  type        = {Bachelor Thesis},
  school      = {Delft University of Technology},
  year        = {2025},
  url         = {https://resolver.tudelft.nl/uuid:7ad41253-902a-4d78-a86a-c7bee94b9e0c}
}

@thesis{Petre2025,
  author      = {I. A. Petre},
  title       = {Performance of Optical Flow Models on Real-World Occluded Regions},
  type        = {Bachelor Thesis},
  school      = {Delft University of Technology},
  year        = {2025},
  url         = {https://resolver.tudelft.nl/uuid:c25761a4-939d-4677-b790-b3fb9bc4bc28}
}

@thesis{Klijnsma2025,
  author      = {J. B. Klijnsma},
  title       = {Real-world Evaluation of Optical Flow on Repetitive Patterns},
  type        = {Bachelor Thesis},
  school      = {Delft University of Technology},
  year        = {2025},
  url         = {https://resolver.tudelft.nl/uuid:dc709377-e116-48eb-91a5-4b6812902034}
}

@article{barron1994performance,
  title={Performance of optical flow techniques},
  author={Barron, John L and Fleet, David J and Beauchemin, Steven S},
  journal={International journal of computer vision},
  volume={12},
  number={1},
  pages={43--77},
  year={1994},
  publisher={Springer}
}

@article{heeger1987model,
  title={Model for the extraction of image flow},
  author={Heeger, David J},
  journal={Journal of the Optical Society of America A},
  volume={4},
  number={8},
  pages={1455--1471},
  year={1987},
  publisher={Optical Society of America}
}

\newpage
\appendix
\begin{center}
\Large\textbf{Appendix}
\end{center}
\section{Reproducing Reported Checkpoint Results (Things)}
\label{sec:reproduce}
\begin{table}[ht]
\centering
\small
\setlength{\tabcolsep}{4pt}
\caption{Paper vs.\ ours (ptlflow). Values are mean errors per model. $\Delta$ = ours $-$ paper.}
\label{tab:paper_vs_ours}
\begin{adjustbox}{width=\columnwidth, totalheight=0.9\textheight, keepaspectratio}
\begin{tabular}{l
        S[table-format=2.3]S[table-format=2.3]S[table-format=+1.3]
        S[table-format=2.3]S[table-format=2.3]S[table-format=+1.3]
        S[table-format=2.3]S[table-format=2.3]S[table-format=+1.3]
        S[table-format=2.3]S[table-format=2.3]S[table-format=+1.3]}
\toprule
 & \multicolumn{3}{c}{Sintel-Clean EPE} & \multicolumn{3}{c}{Sintel-Final EPE}
 & \multicolumn{3}{c}{KITTI EPE} & \multicolumn{3}{c}{KITTI Fl-all} \\
\cmidrule(lr){2-4}\cmidrule(lr){5-7}\cmidrule(lr){8-10}\cmidrule(lr){11-13}
Model & {Paper} & {Ours} & {$\Delta$}
      & {Paper} & {Ours} & {$\Delta$}
      & {Paper} & {Ours} & {$\Delta$}
      & {Paper} & {Ours} & {$\Delta$}\\
\midrule
csflow & 1.420 & 1.426 & 0.006 & 2.600 & 2.618 & 0.018 & 4.690 & 4.649 & -0.041 & 16.500 & 16.034 & -0.466 \\
dicl & 1.940 & 1.967 & 0.027 & 3.770 & 3.732 & -0.038 & 8.700 & 9.659 & 0.959 & 23.600 & 23.718 & 0.118 \\
dip & 1.300 & 1.344 & 0.044 & 2.820 & 3.006 & 0.186 & 4.290 & 4.405 & 0.115 & 13.730 & 13.346 & -0.384 \\
dpflow & 1.020 & 1.020 & 0.000 & 2.260 & 2.262 & 0.002 & 3.370 & 3.387 & 0.017 & 11.100 & 11.094 & -0.006 \\
fastflownet & 2.890 & 2.873 & -0.017 & 4.140 & 4.194 & 0.054 & 12.240 & 12.337 & 0.097 & 33.100 & 33.274 & 0.174 \\
flow1d & 1.980 & 1.976 & -0.004 & 3.270 & 3.240 & -0.030 & 6.690 & 6.582 & -0.108 & 23.000 & 22.365 & -0.635 \\
flow\_anything & 1.060 & 1.160 & 0.100 & 2.220 & 2.652 & 0.432 & 1.670 & 1.919 & 0.249 & 5.270 & 5.568 & 0.298 \\
flowformer & 0.950 & 0.939 & -0.011 & 2.350 & 2.339 & -0.011 & 4.090 & 4.554 & 0.464 & 14.700 & 15.006 & 0.306 \\
flowformer\_pp & 0.900 & 0.900 & -0.000 & 2.300 & 2.296 & -0.004 & 3.930 & 4.335 & 0.405 & 14.100 & 15.018 & 0.918 \\
flownet2 & 2.020 & 2.223 & 0.203 & 3.540 & 3.639 & 0.099 & 10.080 & 10.091 & 0.011 & 30.000 & 30.475 & 0.475 \\
flownetc & 4.310 & 3.109 & -1.201 & 5.870 & 4.608 & -1.262 & \multicolumn{1}{c}{--} & 11.502 & \multicolumn{1}{c}{--} & \multicolumn{1}{c}{--} & 44.189 & \multicolumn{1}{c}{--} \\
flownetcss & 2.100 & 2.210 & 0.110 & 3.230 & 3.551 & 0.321 & 8.940 & 8.906 & -0.034 & 29.770 & 29.314 & -0.456 \\
flownets & 4.500 & 3.983 & -0.517 & 5.450 & 5.264 & -0.186 & \multicolumn{1}{c}{--} & 14.193 & \multicolumn{1}{c}{--} & \multicolumn{1}{c}{--} & 51.171 & \multicolumn{1}{c}{--} \\
flowseek\_m & 1.100 & 1.183 & 0.083 & 2.310 & 2.504 & 0.194 & 3.990 & 4.803 & 0.813 & 12.100 & 13.535 & 1.435 \\
flowseek\_t & 1.130 & 1.112 & -0.018 & 2.480 & 2.495 & 0.015 & 4.060 & 4.133 & 0.073 & 12.200 & 12.726 & 0.526 \\
gma & 1.300 & 1.316 & 0.016 & 2.740 & 2.754 & 0.014 & 4.690 & 4.483 & -0.207 & 17.100 & 16.541 & -0.559 \\
gmflow\_p & 1.500 & 1.495 & -0.005 & 2.960 & 2.955 & -0.005 & \multicolumn{1}{c}{--} & 10.347 & \multicolumn{1}{c}{--} & \multicolumn{1}{c}{--} & 33.545 & \multicolumn{1}{c}{--} \\
gmflow\_p\_sc2 & 1.080 & 1.084 & 0.004 & 2.480 & 2.474 & -0.006 & 7.770 & 6.747 & -1.023 & 23.400 & 21.673 & -1.727 \\
gmflow\_p\_sc2\_ref6 & 0.900 & 0.908 & 0.008 & 2.740 & 2.740 & -0.000 & 5.740 & 4.920 & -0.820 & 17.600 & 16.001 & -1.599 \\
gmflow\_refine & 1.080 & 1.084 & 0.004 & 2.480 & 2.474 & -0.006 & 7.770 & 6.747 & -1.023 & 23.400 & 21.673 & -1.727 \\
gmflownet & 1.140 & 1.242 & 0.102 & 2.710 & 2.691 & -0.019 & 4.240 & 4.181 & -0.059 & 15.400 & 15.018 & -0.382 \\
hd3 & 3.840 & 3.648 & -0.192 & 8.770 & 6.237 & -2.533 & 13.170 & 13.333 & 0.163 & 24.000 & 23.854 & -0.146 \\
irr\_pwc & 2.340 & 1.826 & -0.514 & 3.950 & 3.398 & -0.552 & \multicolumn{1}{c}{--} & 7.855 & \multicolumn{1}{c}{--} & \multicolumn{1}{c}{--} & 24.493 & \multicolumn{1}{c}{--} \\
liteflownet & 2.480 & 2.591 & 0.111 & 4.040 & 4.063 & 0.023 & 10.390 & 10.648 & 0.258 & 28.500 & 30.414 & 1.914 \\
llaflow & 1.380 & 1.272 & -0.108 & 2.610 & 2.733 & 0.123 & \multicolumn{1}{c}{--} & 4.704 & \multicolumn{1}{c}{--} & \multicolumn{1}{c}{--} & 16.333 & \multicolumn{1}{c}{--} \\
maskflownet\_s & 2.330 & 2.535 & 0.205 & 3.720 & 3.885 & 0.165 & \multicolumn{1}{c}{--} & 10.773 & \multicolumn{1}{c}{--} & 23.580 & 32.763 & 9.183 \\
memflow & 0.930 & 1.083 & 0.153 & 2.080 & 2.289 & 0.209 & 3.880 & 4.096 & 0.216 & 13.700 & 14.101 & 0.401 \\
memflow\_t & 0.850 & 0.965 & 0.115 & 2.060 & 2.321 & 0.261 & 3.380 & 4.636 & 1.256 & 12.800 & 14.613 & 1.813 \\
neuflow & 1.660 & 1.660 & 0.000 & 3.130 & 3.126 & -0.004 & 12.400 & 11.111 & -1.289 & 32.500 & 30.919 & -1.581 \\
pwcnet & 2.550 & 2.643 & 0.093 & 3.930 & 4.059 & 0.129 & 10.350 & 10.288 & -0.062 & 33.700 & 33.016 & -0.684 \\
raft & 1.430 & 1.494 & 0.064 & 2.710 & 2.689 & -0.021 & 5.040 & 4.961 & -0.079 & 17.400 & 17.021 & -0.379 \\
raft\_small & 2.210 & 2.125 & -0.085 & 3.350 & 3.247 & -0.103 & 7.510 & 7.460 & -0.050 & 26.900 & 24.652 & -2.248 \\
rapidflow & 1.580 & 1.574 & -0.006 & 2.940 & 2.895 & -0.045 & 5.870 & 5.872 & 0.002 & 17.700 & 17.719 & 0.019 \\
rapidflow\_it3 & 2.030 & 2.026 & -0.004 & 3.360 & 3.355 & -0.005 & \multicolumn{1}{c}{--} & 9.168 & \multicolumn{1}{c}{--} & 25.500 & 25.488 & -0.012 \\
rapidflow\_it6 & 1.690 & 1.685 & -0.005 & 3.000 & 2.996 & -0.004 & \multicolumn{1}{c}{--} & 6.766 & \multicolumn{1}{c}{--} & 19.900 & 19.855 & -0.045 \\
rpknet & 1.120 & 1.116 & -0.004 & 2.450 & 2.448 & -0.002 & 3.780 & 3.787 & 0.007 & 13.000 & 12.993 & -0.007 \\
sea\_raft\_l & 1.190 & 1.226 & 0.036 & 4.110 & 3.444 & -0.666 & 3.620 & 3.738 & 0.118 & 12.900 & 12.379 & -0.521 \\
sea\_raft\_m & 1.210 & 1.273 & 0.063 & 4.040 & 3.852 & -0.188 & 4.290 & 4.288 & -0.002 & 14.200 & 13.932 & -0.268 \\
sea\_raft\_s & 1.270 & 1.283 & 0.013 & 3.740 & 3.727 & -0.013 & 4.430 & 4.435 & 0.005 & 15.100 & 14.699 & -0.401 \\
skflow & 1.220 & 1.197 & -0.023 & 2.460 & 2.426 & -0.034 & 4.470 & 4.253 & -0.217 & 15.500 & 15.136 & -0.364 \\
waft\_dav2\_a1 & 1.000 & 1.009 & 0.009 & 2.150 & 2.171 & 0.021 & 3.100 & 3.106 & 0.006 & 10.300 & 10.110 & -0.190 \\
\bottomrule
\end{tabular}
\end{adjustbox}
\end{table}
\clearpage

\section{FlowFactor results}
\begin{table*}[ht]
  \centering
  \small
\setlength{\tabcolsep}{4pt}
  \caption{Per-category ranking on \textit{FlowFactor}. Mean Fl-all (\%) with bootstrapped 95\% confidence intervals.}
  \begin{adjustbox}{width=\columnwidth}
  \begin{tabular}{lllllllll}
    \toprule
    Rank & Lighting &  & Large disp. &  & Occlusions &  & Repetitive tex. &  \\
    \midrule
    1 & waft\_dav2\_a1 & 1.0 [0.1, 2.3] & ff++ & 4.7 [2.6, 7.1] & waft\_dav2\_a1 & 14.0 [10.1, 18.2] & gmflow\_p\_sc2\_ref6 & 3.0 [1.9, 4.1] \\
    2 & craft & 4.0 [2.3, 5.9] & flowformer & 6.5 [4.2, 9.0] & dpflow & 15.2 [11.3, 19.4] & gmflownet & 3.8 [2.2, 5.6] \\
    3 & rpknet & 5.0 [3.0, 7.2] & gf\_ref & 6.6 [4.1, 9.3] & memflow & 18.2 [13.8, 22.9] & flowformer & 3.9 [2.3, 5.7] \\
    4 & rapidflow & 6.7 [4.6, 9.0] & gfpsc2 & 6.6 [4.1, 9.4] & ff++ & 18.7 [14.4, 23.4] & ff++ & 4.1 [2.4, 6.0] \\
    5 & flow1d & 6.8 [4.4, 9.4] & gmflow\_p\_sc2\_ref6 & 6.9 [4.3, 9.8] & rpknet & 18.7 [14.3, 23.4] & gf\_ref & 4.8 [3.2, 6.6] \\
    6 & raft & 6.9 [4.6, 9.6] & gf & 8.9 [6.2, 11.8] & flowformer & 18.8 [14.3, 23.5] & gfpsc2 & 4.8 [3.2, 6.6] \\
    7 & flownet2 & 7.0 [5.1, 9.0] & gf\_p & 8.9 [6.3, 11.8] & llaflow & 19.3 [14.7, 24.1] & craft & 5.1 [3.0, 7.4] \\
    8 & llaflow & 7.0 [4.7, 9.6] & gmflownet & 10.1 [7.0, 13.3] & memflow\_t & 19.5 [15.0, 24.3] & llaflow & 5.3 [3.2, 7.6] \\
    9 & gma & 7.4 [4.7, 10.2] & memflow\_t & 14.4 [10.6, 18.5] & csflow & 19.6 [15.1, 24.3] & skflow & 5.4 [3.2, 7.8] \\
    10 & gmflownet & 7.4 [5.0, 10.0] & flow1d & 15.0 [11.2, 18.8] & gmflow\_p\_sc2\_ref6 & 19.8 [15.5, 24.2] & memflow & 5.7 [3.5, 8.1] \\
    11 & skflow & 7.5 [5.2, 9.9] & memflow & 16.6 [12.2, 21.0] & skflow & 20.3 [15.7, 25.1] & waft\_dav2\_a1 & 6.0 [3.9, 8.4] \\
    12 & dpflow & 7.8 [5.6, 10.2] & neuflow & 18.2 [14.2, 22.4] & sea\_raft\_l & 20.9 [16.1, 25.7] & dpflow & 6.4 [4.2, 8.9] \\
    13 & memflow & 8.4 [6.0, 11.2] & craft & 20.1 [15.5, 25.0] & craft & 21.8 [17.0, 26.7] & gma & 6.7 [4.3, 9.4] \\
    14 & csflow & 8.8 [6.2, 11.7] & neuflow2 & 20.1 [16.1, 24.4] & gmflownet & 22.1 [17.4, 26.9] & dip & 7.0 [4.6, 9.8] \\
    15 & rapidflow\_it6 & 9.4 [7.0, 12.2] & skflow & 25.7 [20.6, 31.0] & neuflow2 & 22.8 [18.1, 27.6] & rpknet & 7.0 [4.6, 9.6] \\
    16 & flowseek\_m & 9.5 [7.3, 11.9] & llaflow & 26.0 [20.8, 31.2] & dip & 22.9 [17.9, 27.8] & raft & 7.1 [4.7, 9.7] \\
    17 & irr\_pwc & 10.5 [7.8, 13.5] & rpknet & 26.7 [21.7, 31.9] & flow1d & 23.4 [18.6, 28.6] & gf & 7.2 [5.1, 9.5] \\
    18 & flowformer & 11.0 [8.4, 13.8] & csflow & 28.1 [23.0, 33.3] & raft & 23.9 [19.1, 29.0] & gf\_p & 7.2 [5.1, 9.5] \\
    19 & gmflow\_p\_sc2\_ref6 & 11.7 [8.8, 15.0] & raft & 28.2 [23.2, 33.3] & sea\_raft\_m & 24.5 [19.8, 29.4] & neuflow & 7.4 [5.4, 9.6] \\
    20 & dicl & 11.8 [9.0, 14.7] & sea\_raft\_l & 29.6 [24.6, 34.7] & gma & 25.3 [20.2, 30.5] & neuflow2 & 8.1 [5.9, 10.5] \\
    21 & ff++ & 12.5 [9.7, 15.5] & waft\_dav2\_a1 & 29.7 [24.4, 35.1] & flowseek\_t & 25.7 [20.8, 30.6] & csflow & 8.1 [5.4, 11.1] \\
    22 & rapidflow\_it3 & 12.6 [9.7, 15.5] & gma & 31.1 [25.8, 36.5] & flowseek\_m & 26.1 [21.3, 31.2] & memflow\_t & 8.1 [5.5, 11.2] \\
    23 & dip & 12.7 [9.4, 16.3] & dpflow & 32.5 [27.2, 37.9] & gf\_ref & 29.7 [24.6, 34.8] & flow1d & 10.1 [7.2, 13.2] \\
    24 & neuflow2 & 14.0 [10.9, 17.1] & irr\_pwc & 33.1 [27.6, 38.9] & gfpsc2 & 29.7 [24.6, 34.8] & rapidflow & 10.3 [7.1, 13.7] \\
    25 & gf & 14.1 [10.6, 17.9] & rapidflow & 36.2 [30.6, 41.8] & sea\_raft\_s & 30.3 [25.1, 35.6] & irr\_pwc & 10.6 [7.7, 13.8] \\
    26 & gf\_p & 14.1 [10.6, 17.8] & flowseek\_t & 36.6 [31.5, 42.0] & rapidflow & 32.4 [26.8, 37.8] & rapidflow\_it6 & 12.7 [9.2, 16.5] \\
    27 & gf\_ref & 14.5 [11.0, 18.2] & dicl & 37.4 [31.9, 43.0] & irr\_pwc & 33.0 [27.6, 38.5] & flowseek\_t & 13.9 [10.5, 17.5] \\
    28 & gfpsc2 & 14.5 [11.0, 18.3] & sea\_raft\_m & 39.5 [34.5, 44.6] & gf & 35.7 [30.2, 41.0] & sea\_raft\_l & 16.9 [13.0, 21.1] \\
    29 & memflow\_t & 15.2 [11.8, 18.9] & rapidflow\_it6 & 42.5 [36.6, 48.5] & gf\_p & 35.7 [30.3, 41.1] & sea\_raft\_m & 17.1 [13.3, 21.0] \\
    30 & sea\_raft\_l & 15.5 [11.9, 19.6] & sea\_raft\_s & 42.5 [37.0, 47.9] & rapidflow\_it6 & 37.1 [31.6, 42.5] & sea\_raft\_s & 18.8 [14.7, 22.8] \\
    31 & flownetcss & 15.7 [12.2, 19.2] & dip & 43.2 [37.4, 49.0] & neuflow & 37.1 [31.8, 42.4] & flowseek\_m & 19.1 [15.1, 23.3] \\
    32 & flowseek\_t & 16.0 [12.8, 19.3] & flowseek\_m & 43.8 [38.2, 49.3] & flownet2 & 45.5 [39.9, 51.4] & fastflownet & 22.2 [18.1, 26.5] \\
    33 & sea\_raft\_m & 17.7 [14.1, 21.5] & flownetcss & 46.2 [40.3, 52.1] & flownetcss & 46.2 [40.5, 51.8] & dicl & 22.6 [17.8, 27.6] \\
    34 & pwcnet & 18.2 [14.7, 21.9] & maskflownet\_s & 46.2 [40.7, 51.8] & pwcnet & 47.0 [41.6, 52.6] & pwcnet & 24.0 [19.8, 28.5] \\
    35 & liteflownet & 20.2 [16.6, 24.1] & pwcnet & 46.3 [40.5, 52.1] & maskflownet\_s & 48.6 [42.9, 54.2] & maskflownet\_s & 24.2 [19.7, 28.6] \\
    36 & fastflownet & 20.8 [17.1, 24.7] & rapidflow\_it3 & 47.2 [41.4, 53.2] & dicl & 49.6 [44.2, 55.1] & rapidflow\_it3 & 25.1 [20.6, 29.9] \\
    37 & flownetc & 21.5 [17.7, 25.4] & flownet2 & 49.0 [43.2, 54.7] & liteflownet & 55.3 [49.7, 60.7] & flownetcss & 26.1 [21.4, 30.8] \\
    38 & neuflow & 21.9 [18.1, 25.9] & fastflownet & 50.7 [45.4, 56.2] & rapidflow\_it3 & 56.3 [50.8, 61.8] & liteflownet & 29.9 [25.2, 34.8] \\
    39 & sea\_raft\_s & 22.2 [18.2, 26.2] & liteflownet & 51.2 [45.4, 57.3] & fastflownet & 58.6 [53.3, 63.9] & flownet2 & 32.0 [27.0, 37.0] \\
    40 & maskflownet\_s & 25.6 [21.5, 29.8] & flownetc & 60.9 [55.4, 66.2] & flownetc & 71.1 [66.4, 75.8] & flownetc & 47.0 [41.9, 51.9] \\
    41 & flownets & 32.6 [28.3, 37.0] & flownets & 67.7 [62.2, 73.0] & flownets & 82.6 [78.4, 86.5] & flownets & 66.9 [62.1, 71.5] \\
    \bottomrule
  \end{tabular}
  \end{adjustbox}
  \label{tab:flowfactor_ranking}
\end{table*}

\clearpage

\section{Results Real-world benchmark}
\label{sec:realworld}
\begin{table}[ht]
\centering
\caption{Fl-error scores and relative rankings across three datasets. Models (things) are sorted by their overall average rank.}
\label{tab:benchmark_results}
\begin{adjustbox}{width=\columnwidth}
\begin{tabular}{lS[table-format=2.1]S[table-format=2.1]S[table-format=2.1]S[table-format=2.1]S[table-format=2.1]S[table-format=2.1]S[table-format=2.1]}
\toprule
  & \multicolumn{2}{c}{FlowFactor} & \multicolumn{2}{c}{Slow Flow\cite{janai2017slow}} & \multicolumn{2}{c}{TAP-Flow\cite{doersch2022tap}} &   \\
\cmidrule(lr){2-3} \cmidrule(lr){4-5} \cmidrule(lr){6-7}
Model & {Score} & {Rank} & {Score} & {Rank} & {Score} & {Rank} & {Avg.\ Rank} \\
\midrule
rpknet & 14.4 & 11.0 & 26.8 & 2.0 & 31.6 & 2.0 & 5.0 \\
flowformer\_pp & 10.1 & 1.0 & 28.3 & 6.0 & 35.1 & 12.0 & 6.3 \\
flowformer & 10.1 & 2.0 & 28.5 & 8.0 & 34.9 & 11.0 & 7.0 \\
gmflownet & 10.9 & 4.0 & 28.4 & 7.0 & 35.4 & 14.0 & 8.3 \\
rapidflow & 21.5 & 22.0 & 26.7 & 1.0 & 32.4 & 4.0 & 9.0 \\
memflow & 12.3 & 5.0 & 28.9 & 12.0 & 34.9 & 10.0 & 9.0 \\
waft\_dav2\_a1 & 12.7 & 6.0 & 28.1 & 4.0 & 37.8 & 18.0 & 9.3 \\
dpflow & 15.5 & 14.0 & 29.1 & 14.0 & 31.4 & 1.0 & 9.7 \\
skflow & 14.8 & 13.0 & 28.3 & 5.0 & 35.7 & 15.0 & 11.0 \\
llaflow & 14.5 & 12.0 & 28.7 & 10.0 & 35.1 & 13.0 & 11.7 \\
rapidflow\_it6 & 25.5 & 28.0 & 27.2 & 3.0 & 33.3 & 8.0 & 13.0 \\
csflow & 16.2 & 15.0 & 29.2 & 17.0 & 33.2 & 7.0 & 13.0 \\
gma & 17.7 & 19.0 & 28.9 & 11.0 & 36.1 & 16.0 & 15.3 \\
gmflow\_p\_sc2\_ref6 & 10.4 & 3.0 & 30.4 & 24.0 & 38.0 & 19.0 & 15.3 \\
raft & 16.6 & 18.0 & 29.4 & 19.0 & 33.4 & 9.0 & 15.3 \\
flowseek\_t & 23.1 & 25.0 & 30.3 & 23.0 & 32.9 & 6.0 & 18.0 \\
gmflow\_refine & 14.0 & 8.5 & 29.6 & 20.5 & 41.4 & 25.5 & 18.2 \\
gmflow\_p\_sc2 & 14.0 & 8.5 & 29.6 & 20.5 & 41.4 & 25.5 & 18.2 \\
dip & 21.5 & 23.0 & 31.1 & 29.0 & 31.8 & 3.0 & 18.3 \\
flowseek\_m & 24.7 & 26.0 & 30.5 & 25.0 & 32.8 & 5.0 & 18.7 \\
memflow\_t & 14.4 & 10.0 & 30.8 & 27.0 & 38.2 & 20.0 & 19.0 \\
neuflow & 21.3 & 21.0 & 28.7 & 9.0 & 44.1 & 29.0 & 19.7 \\
irr\_pwc & 21.8 & 24.0 & 29.2 & 16.0 & 39.3 & 21.0 & 20.3 \\
flownet2 & 33.4 & 31.0 & 29.0 & 13.0 & 40.4 & 23.0 & 22.3 \\
flow1d & 13.8 & 7.0 & 31.0 & 28.0 & 45.7 & 33.0 & 22.7 \\
rapidflow\_it3 & 35.3 & 34.0 & 29.3 & 18.0 & 37.1 & 17.0 & 23.0 \\
dicl & 30.4 & 30.0 & 30.1 & 22.0 & 39.6 & 22.0 & 24.7 \\
flownetcss & 33.6 & 32.0 & 30.7 & 26.0 & 41.8 & 27.0 & 28.3 \\
gmflow\_p & 16.6 & 16.5 & 31.5 & 32.5 & 58.0 & 37.0 & 28.7 \\
sea\_raft\_s & 28.5 & 29.0 & 32.9 & 35.0 & 40.7 & 24.0 & 29.3 \\
sea\_raft\_l & 20.8 & 20.0 & 36.2 & 38.0 & 46.1 & 35.0 & 31.0 \\
sea\_raft\_m & 24.8 & 27.0 & 35.6 & 37.0 & 44.9 & 32.0 & 32.0 \\
pwcnet & 34.0 & 33.0 & 31.7 & 34.0 & 44.8 & 31.0 & 32.7 \\
fastflownet & 38.2 & 35.0 & 31.2 & 30.0 & 45.9 & 34.0 & 33.0 \\
liteflownet & 39.2 & 36.0 & 33.3 & 36.0 & 44.1 & 30.0 & 34.0 \\
flownetc & 50.2 & 37.0 & 39.5 & 39.0 & 55.3 & 36.0 & 37.3 \\
flownets & 62.5 & 38.0 & 40.0 & 40.0 & 68.3 & 38.0 & 38.7 \\
\bottomrule
\end{tabular}
\end{adjustbox}
\end{table}

\clearpage

\section{Examples Slow Flow \cite{janai2017slow} \&  TAP-Flow \cite{doersch2022tap}}
\label{sec:examples}
\begin{figure}[ht]
    \centering
    \begin{subfigure}{0.48\linewidth}
        \centering
        \includegraphics[width=\linewidth]{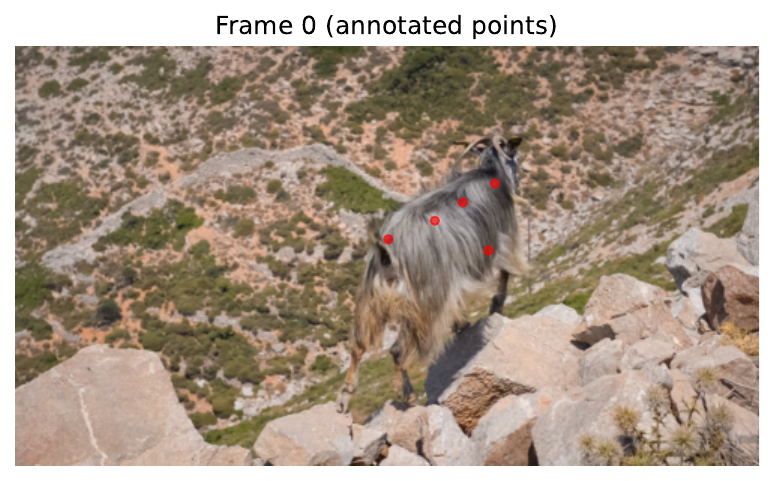}
        \label{fig:disp10}
    \end{subfigure}
    \hfill
    \begin{subfigure}{0.48\linewidth}
        \centering
        \includegraphics[width=\linewidth]{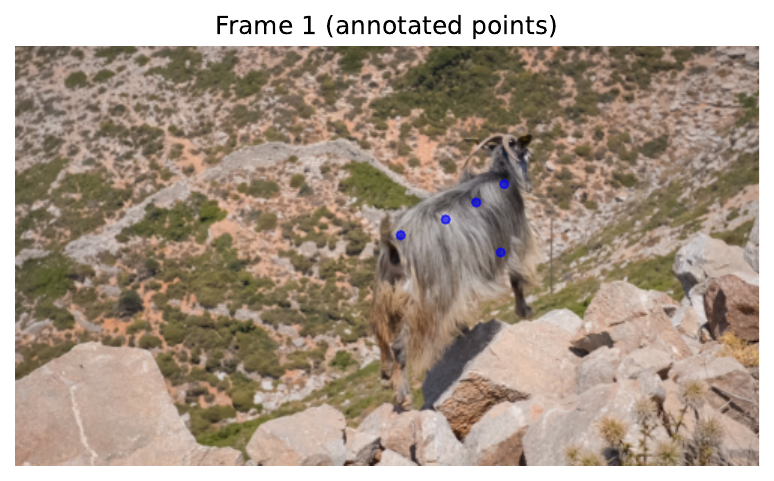}
        \label{fig:disp11}
    \end{subfigure}
    \caption{Example of annotated pixels in the TAP-Flow dataset \cite{doersch2022tap}. Just like the FlowFactor dataset the points were randomly chosen by the annotator. This image shows the ability to accurately annotate non-rigid motion.}
    \label{fig:ex-tapflow}
\end{figure}

\begin{figure}[ht]
    \centering
    \begin{subfigure}{0.48\linewidth}
        \centering
        \includegraphics[width=\linewidth]{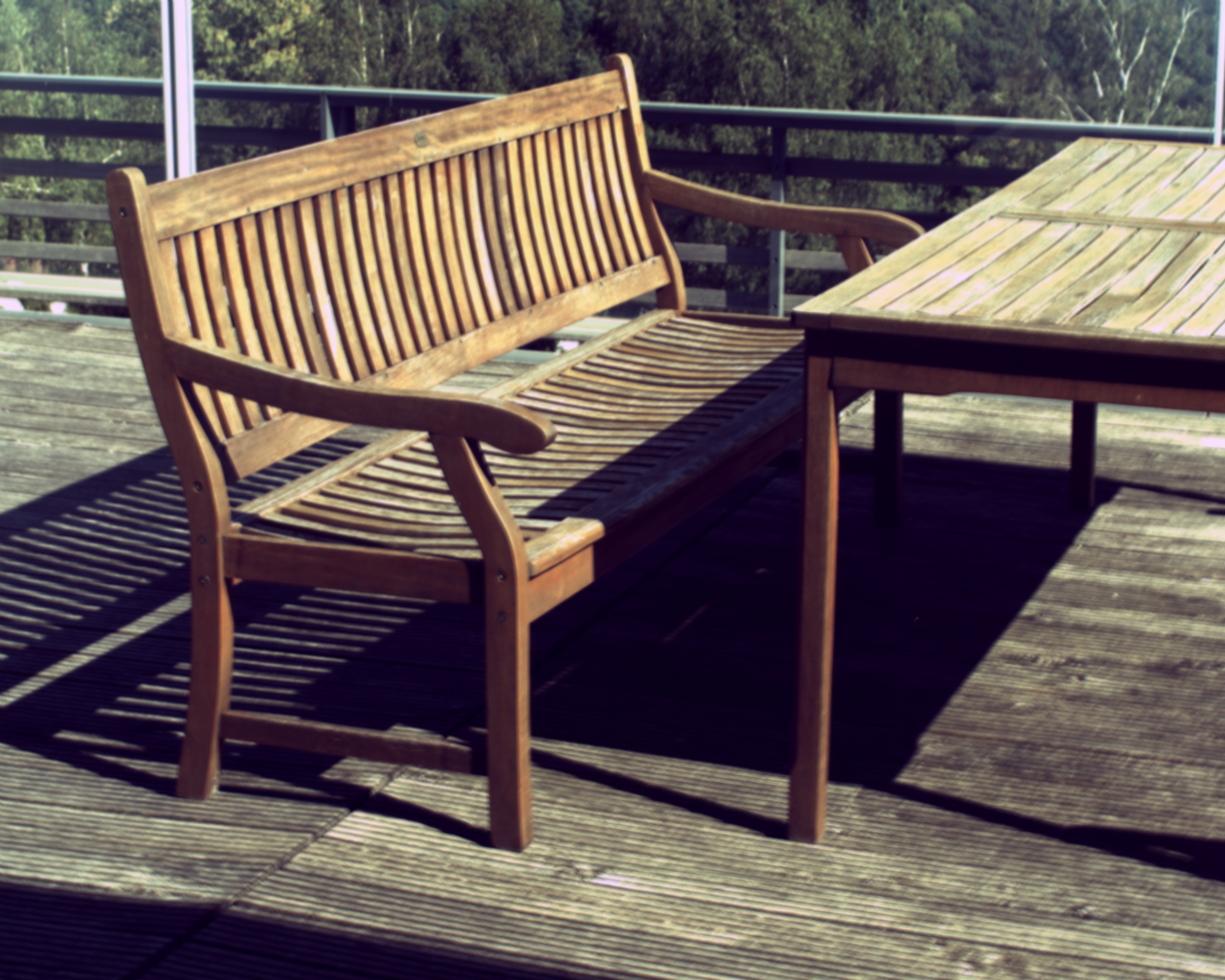}
        \label{fig:disp10}
    \end{subfigure}
    \hfill
    \begin{subfigure}{0.48\linewidth}
        \centering
        \includegraphics[width=\linewidth]{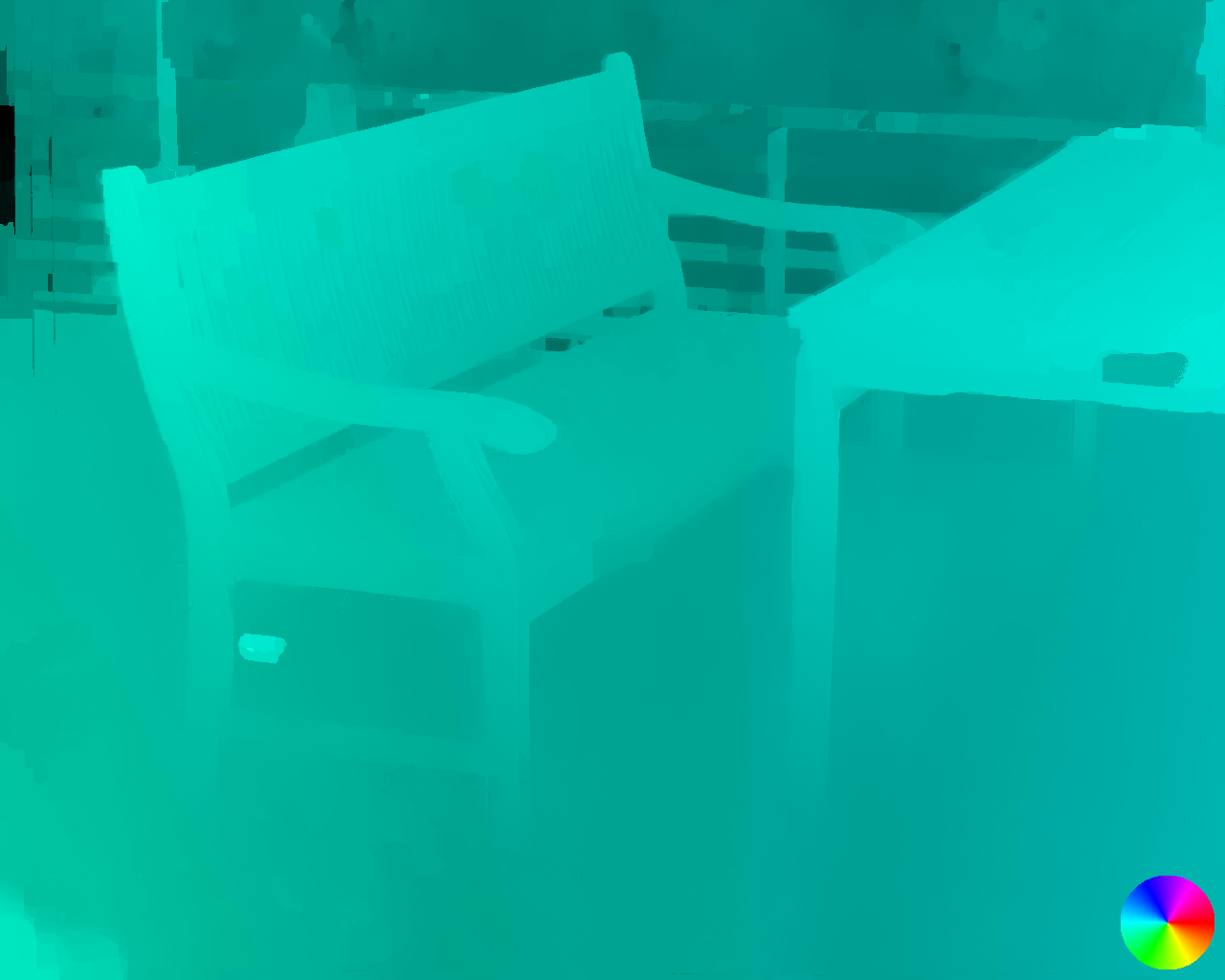}
        \label{fig:disp11}
    \end{subfigure}
    \caption{Example of annotated pixels in the Slow Flow dataset \cite{janai2017slow}. As opposed to the other datasets in our real-world benchmark this utilises a dense flow field. This image showcases the ability of Slow Flow \cite{janai2017slow} to account for occlusions as opposed to the TAP-Flow \cite{doersch2022tap} dataset.}
    \label{fig:slow-ex}
\end{figure}

\clearpage

\section{Annotation tool}
\label{sec:annotation}
\begin{figure}[]
    \centering
    \begin{subfigure}{\linewidth}
        \centering
        \includegraphics[width=\linewidth]{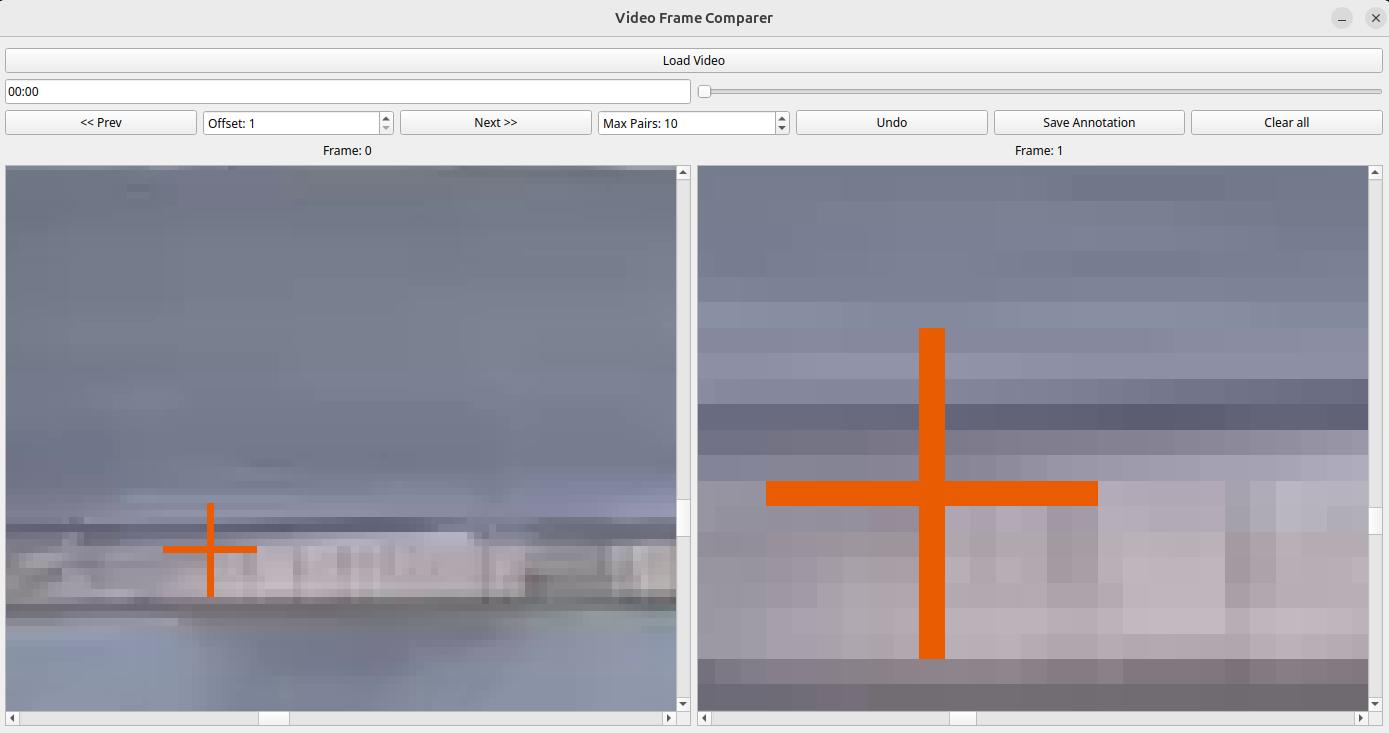}
        \label{fig:disp10}
    \end{subfigure}

    \caption{A screenshot of the annotation tool used to annotate the FlowFactor dataset. It allows for setting a maximum of frame pairs, and the use of an offset to skip a certain amount of frames.}
    \label{fig:slow-ex}
\end{figure}
\clearpage
\section{Training dataset size vs. real-world performance}
\begin{figure}[h]
  \centering
  \includegraphics[width=0.64\linewidth]{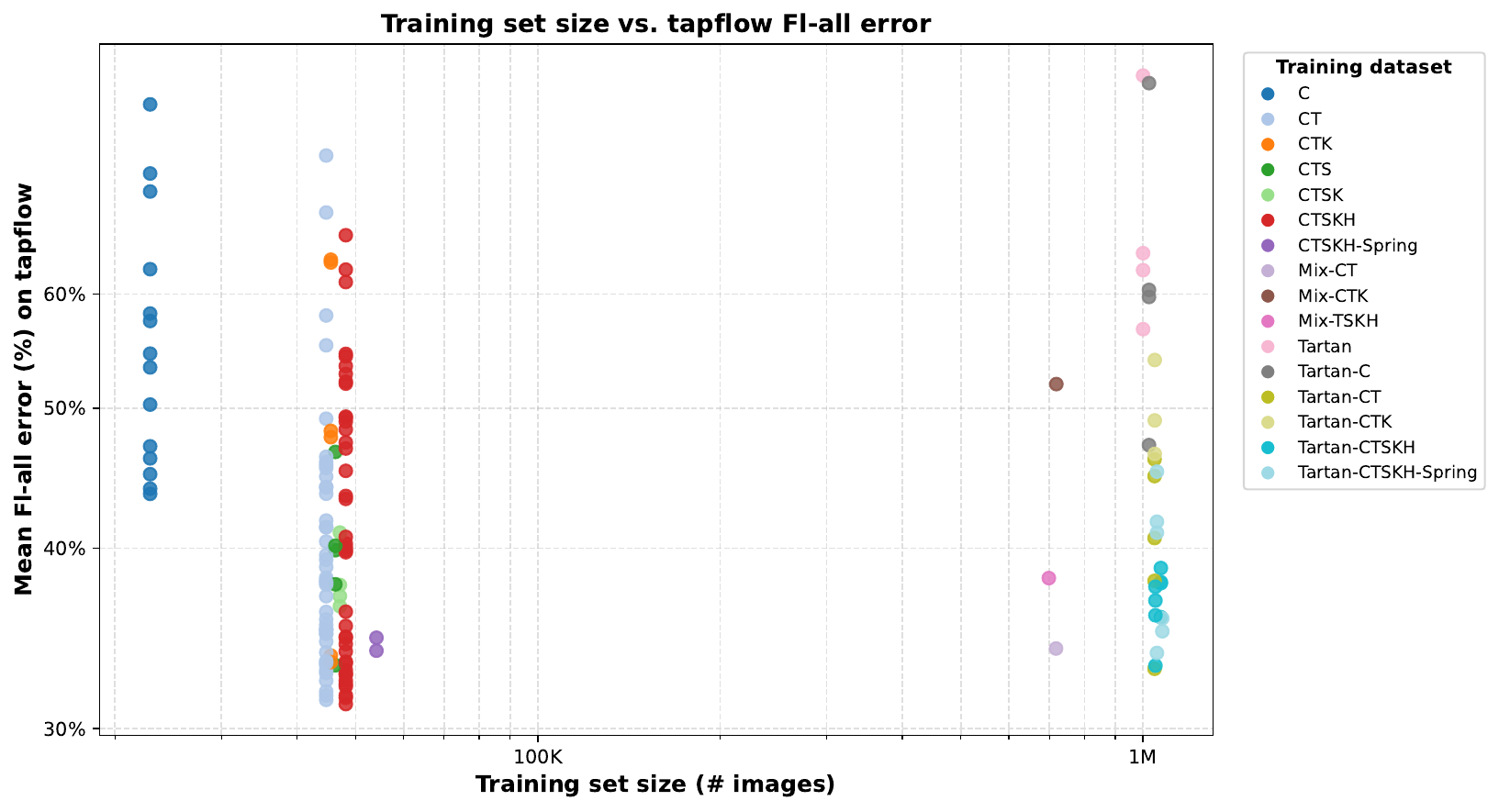}%
  \caption{\textbf{Size training dataset vs. accuracy on TAP-Flow.}
  Here we plot for all models all possible checkpoints. For every checkpoint we plot the accompanying training dataset size and accuracy on the real-world dataset TAP-Flow. Increasing the pre-training dataset size with more synthetic data, does not necessarily improve accuracy on the TAP-Flow dataset, compared to C+T training.}
  \label{fig:train_size_tapflow}
\end{figure}
\begin{figure}[]
  \centering
  \includegraphics[width=0.64\linewidth]{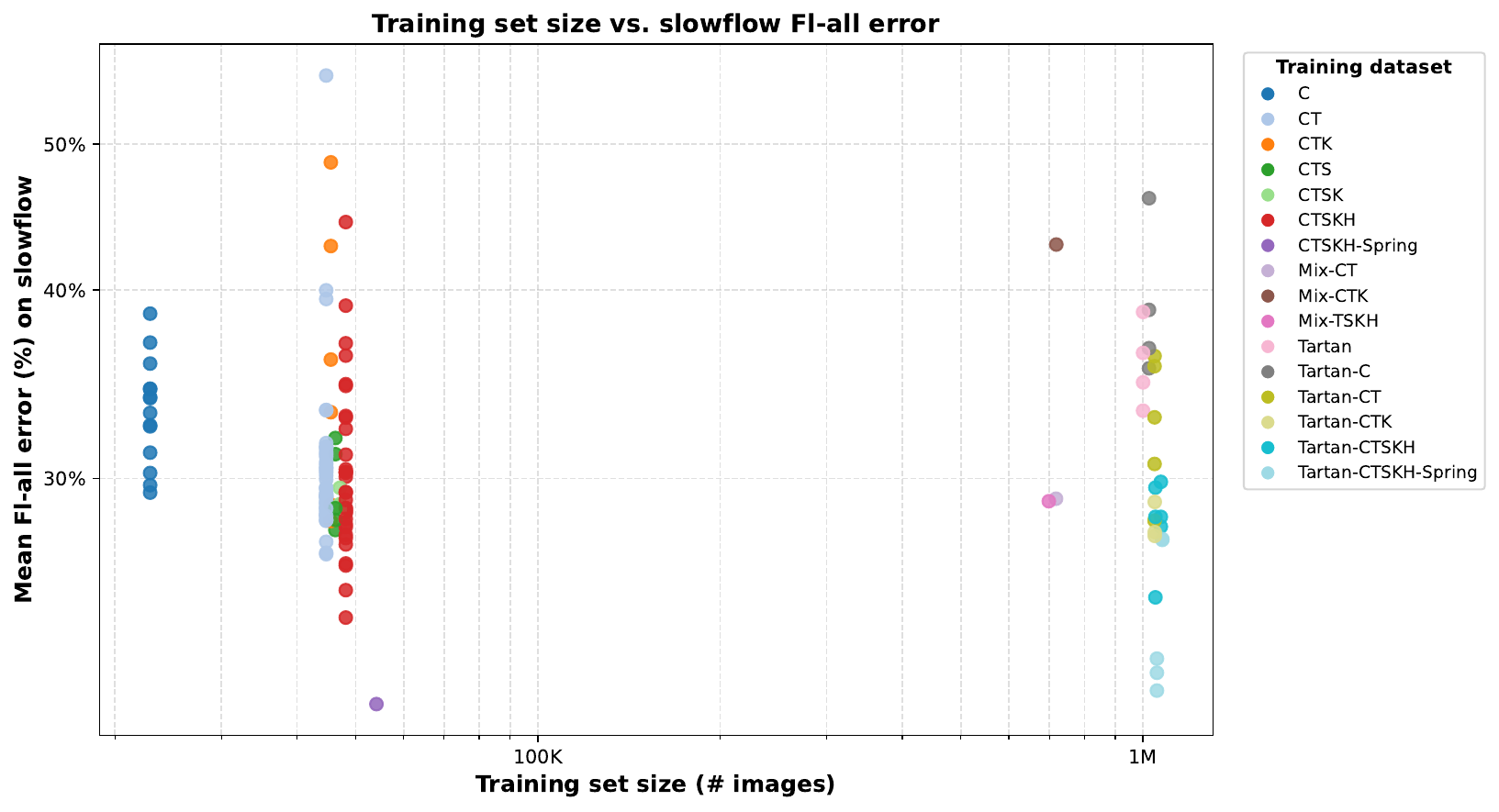}%
  \caption{\textbf{Size training dataset vs. accuracy on Slow Flow.}
  Here we plot for all models all possible checkpoints. For every checkpoint we plot the accompanying training dataset size and accuracy on the real-world dataset Slow Flow. Increasing the pre-training dataset size with more synthetic data, does not necessarily improve accuracy on the Slow Flow dataset, compared to C+T training.}
  \label{fig:train_size_tapflow}
\end{figure}

\end{document}